\documentclass[11pt,a4paper]{article}
\usepackage[margin=0.75in]{geometry}
\usepackage[utf8]{inputenc}
\usepackage{cmap}
\usepackage[T1]{fontenc}
\usepackage{lmodern}
\usepackage{natbib}
\usepackage[colorlinks=true]{hyperref}
\usepackage{amssymb, amsthm, amsmath, amsfonts}
\usepackage{algorithm}
\usepackage{algpseudocode}
\usepackage{authblk, dsfont, mathrsfs, color, bm, enumitem, mathtools}
\usepackage{subfigure, graphicx, transparent}
\usepackage{dirtytalk}
\graphicspath{{figures/}}
\usepackage[dvipsnames]{xcolor}
\usepackage{comment}
\usepackage{tabularx}
\usepackage{booktabs}
\hypersetup{colorlinks, linkcolor={red!80!black}, citecolor={blue!80!black}, urlcolor={ForestGreen}}
\usepackage{multirow}
\usepackage[capitalize,noabbrev]{cleveref}
\usepackage{hyperref}

\theoremstyle{plain}
\newtheorem{theorem}{Theorem}[section]
\newtheorem{proposition}[theorem]{Proposition}
\newtheorem{lemma}[theorem]{Lemma}

\theoremstyle{definition}

\newtheorem{remark}[theorem]{Remark}

\DeclareUnicodeCharacter{211D}{\ensuremath{\mathbb R}}

\newcommand{\cov}{{\mathop {\rm cov{}}}}

\newcommand{\Perp}{\perp\!\!\!\perp}
\newcommand{\rb}{\mathbb{R}}

\def \ds{\displaystyle}

\def \S{  { \Sigma} }

\def \E{{\mathbb E}}
\def \P{{\mathbb P}}

\newcommand{\bivec}[2]{\left(   \begin{array}{c} {#1} \\ {#2} \end{array}   \right)}
\newcommand{\bimat}[4]{\left(    \begin{array}{c@{\hspace{2mm}}c} {#1} &  {#2} \\ {#3} &  {#4} \end{array}     \right)}

\usepackage{pifont}

\usepackage[page]{appendix}

\makeatletter
\let\oldappendix\appendices

\renewcommand{\appendices}{%
  \clearpage
  \RestoreAddContentsLine 
  \renewcommand{\thesection}{\Roman{section}}
  \let\tf@toc\tf@app
  \addtocontents{app}{\protect\setcounter{tocdepth}{2}}
  \immediate\write\@auxout{%
    \string\let\string\tf@toc\string\tf@app^^J
  }
  \oldappendix
}%

\newcommand{\listofappendices}{%
  \begingroup
  \renewcommand{\contentsname}{\appendixtocname}
  \let\@oldstarttoc\@starttoc
  \def\@starttoc##1{\@oldstarttoc{app}}
  \tableofcontents
  \endgroup
}

\makeatother

\makeatletter
\newcommand{\RestoreAddContentsLine}{%
  \ifcsname hyper@anchor\endcsname
    \def\addcontentsline##1##2##3{%
      \addtocontents{##1}{%
        \protect\contentsline{##2}{##3}{\thepage}{\@currentHref}%
      }%
    }%
  \else
    \def\addcontentsline##1##2##3{%
      \addtocontents{##1}{%
        \protect\contentsline{##2}{##3}{\thepage}%
      }%
    }%
  \fi
}
\makeatother

\DeclarePairedDelimiterX{\inner}[2]{\langle}{\rangle}{#1, #2}

\title{Multivariate Standardized Residuals for Conformal Prediction}

\author[1]{Sacha Braun}
\author[1]{Eugène Berta}
\author[$1, 2$]{Michael I. Jordan}
\author[1]{Francis Bach}
\affil[1]{Sierra team, Inria Paris, France \protect\\
\texttt{\{sacha.braun, eugene.berta, francis.bach\}@inria.fr}}
\affil[2]{Departments of EECS and Statistics, UC Berkeley, USA \protect\\ \texttt{jordan@cs.berkeley.edu}}

\date{}

\begin{document}

\maketitle


\begin{abstract}

While split conformal prediction guarantees marginal coverage, approaching the stronger property of conditional coverage is essential for reliable uncertainty quantification.
Naive conformal scores, however, suffer from poor conditional coverage in heteroskedastic settings.
In univariate regression, this is commonly addressed by normalizing non-conformity scores using an estimated local score variance.
In this work, we propose a natural extension of this normalization to the multivariate setting, effectively whitening the residuals to decouple output correlations and standardize local variance.
Furthermore, we derive a sufficient condition characterizing a broad class of distributions for which standardized residuals yield asymptotic conditional coverage.
We demonstrate that using the Mahalanobis distance induced by a learned local covariance as a non-conformity score provides a closed-form, computationally efficient mechanism for capturing inter-output correlations and heteroskedasticity, avoiding the expensive sampling required by previous methods based on cumulative distribution functions.
This structure unlocks several practical extensions, including the handling of missing output values, the refinement of conformal sets when partial information is revealed, and the construction of valid conformal sets for transformations of the output.
Finally, we provide extensive empirical evidence on both synthetic and real-world datasets showing that our approach yields conformal sets that improve upon the conditional coverage of existing multivariate baselines.

\end{abstract}


\section{Introduction}
\label{sec:intro}

Conformal prediction provides a powerful, model-agnostic framework for constructing predictive sets with guaranteed finite-sample marginal coverage \citep{vovk2005algorithmic, shafer2008tutorial}.
Its flexibility makes it applicable to a wide range of predictive models, including black-box predictors.
A significant limitation, however, lies in achieving the stronger guarantee of conditional coverage, where the validity holds conditionally on the observed input. This is known to be impossible without additional assumptions \citep{vovk2012conditional, lei2014distribution, foygel2021limits}, and the limitation is particularly apparent in methods that produce fixed-size prediction sets for every input.
Such methods cannot adequately capture heteroskedasticity, where prediction uncertainty is not uniform. Consequently, a primary focus of current research is the pursuit of improved empirical conditional coverage.

One reasonable path forward is to make use of quantile regression, which estimates quantiles of the conditional distribution, and therefore is adaptive to data heteroskedasticity \citep{romano2019conformalized, angelopoulos2021gentle}.
An alternative strategy, often referred to as ``standardized'' or ``studentized'' residuals in the literature, normalizes the error using a local estimate of the variance \citep{papadopoulos2008normalized, lei2018distribution}.
It is widely adopted because it is not tied to a single coverage level.
This enables more flexible and informative uncertainty quantification.
However, these two approaches are limited to the univariate setting.

An alternative approach is to estimate the entire conditional density of the outcome, rather than focusing on specific quantiles, and to use this density to construct conformal sets \citep{izbicki2022cd, sampson2024flexible, dheur2025unified}. While these methods yield valid conditional coverage when the underlying distribution is accurately modeled, they require tackling multivariate conditional density estimation, which is notoriously challenging \citep{scott2011multivariate}. Furthermore, even with an accurate density estimate, the resulting conformal sets typically lack closed-form solutions and must be approximated via sampling methods. More fundamentally, the primary appeal of the conformal prediction framework lies in its distribution-free nature. Requiring full conditional density estimation is often overly demanding in this context, motivating the need for intermediate strategies.

In this paper, we propose a novel framework that naturally extends standardized residuals to the multivariate setting, by learning a local covariance matrix to rescale the residual errors. We provide theoretical guarantees by establishing a sufficient condition under which our method achieves asymptotic conditional coverage for a broad class of distributions.
Through extensive synthetic and real data experiments, we demonstrate that our method approximates conditional coverage more closely than existing multivariate regression baselines. We also show that it enables several extensions to standard conformal prediction.

\begin{figure}[h!]
    \center
    \vspace{-3mm}
    \subfigure{\includegraphics[width=0.47\columnwidth]{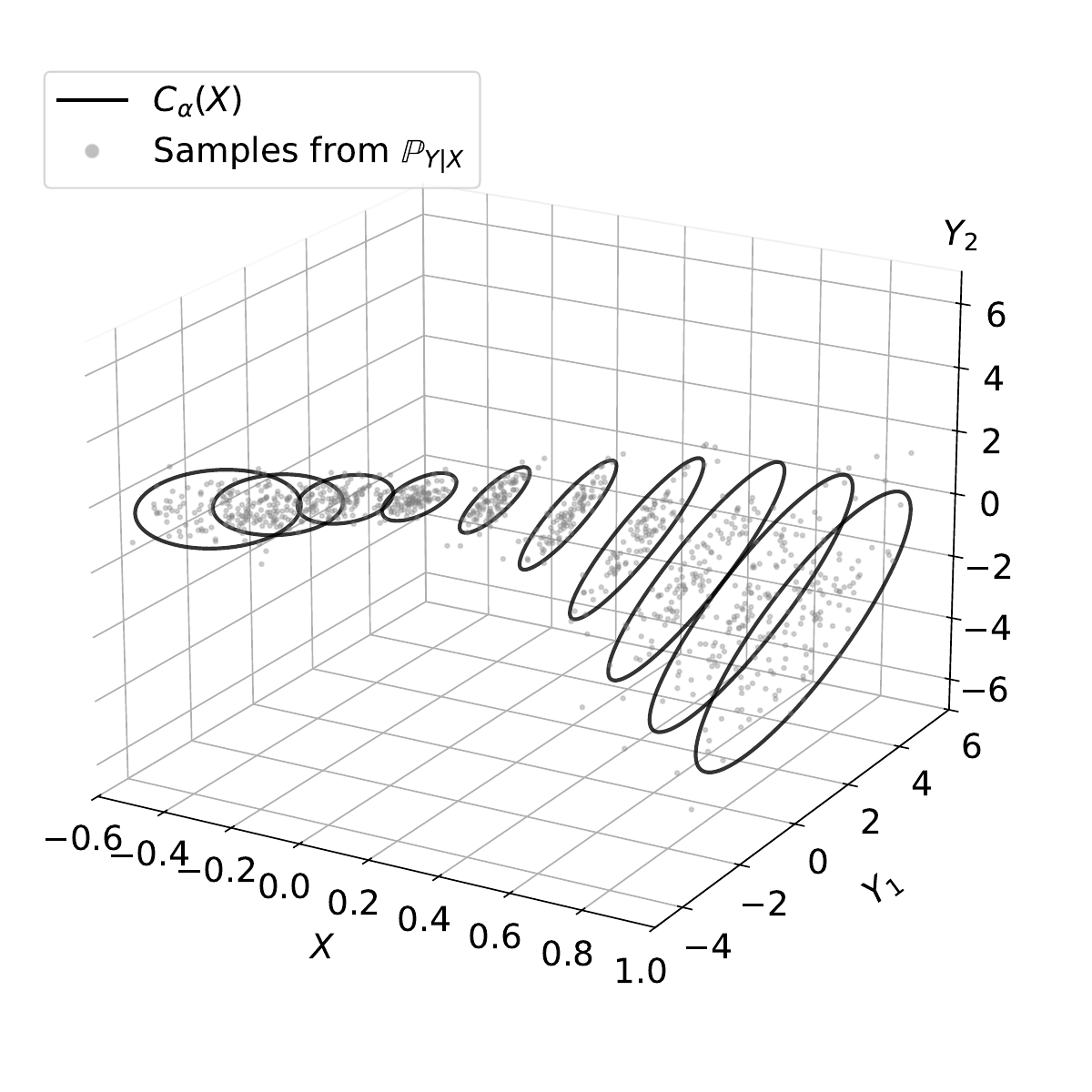}} ~
    \subfigure{\includegraphics[width=0.47\columnwidth]{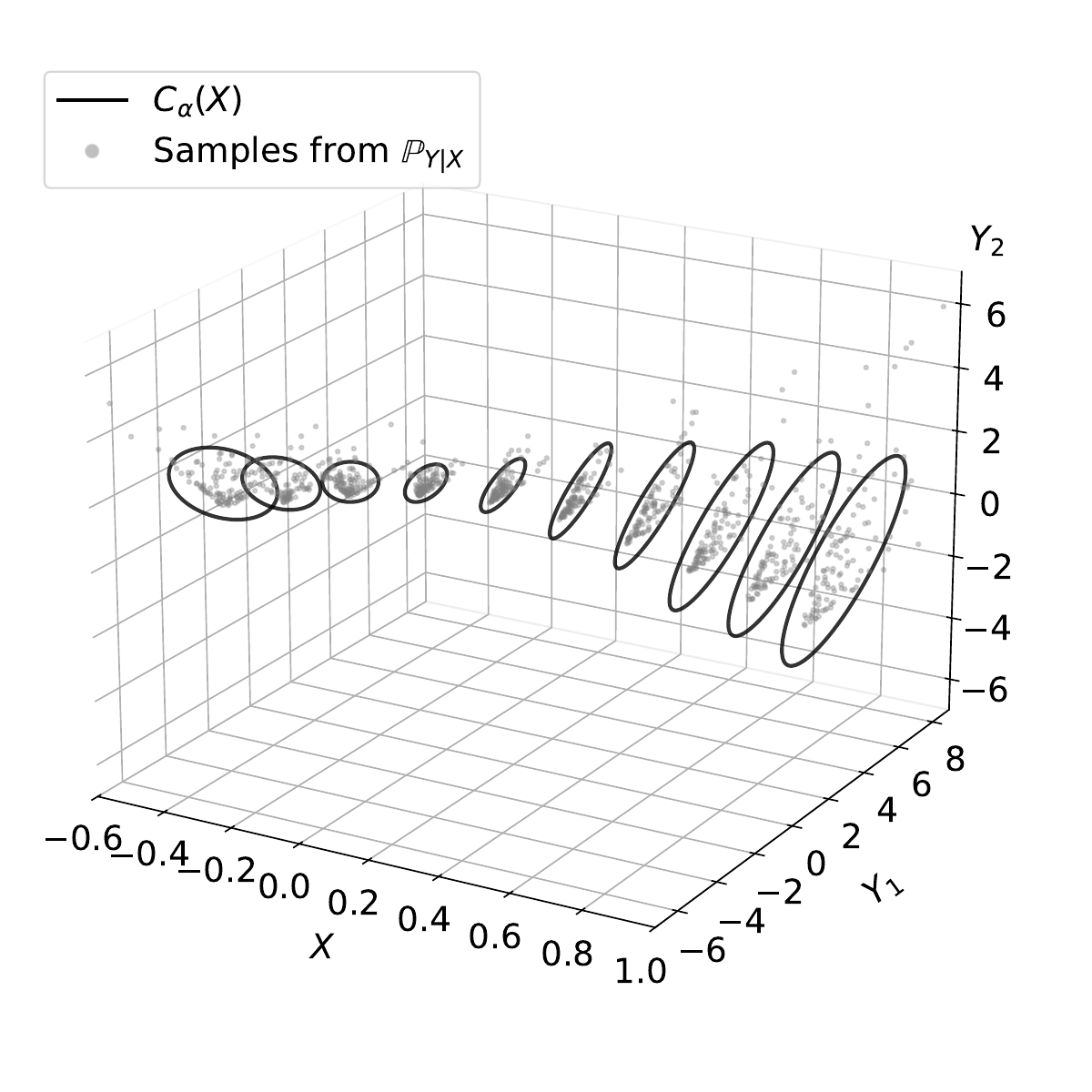}} 
    \vspace{-6mm}
    \caption{Conformal sets obtained with multivariate standardized residuals for different values of the feature variable $X\in[-0.6, 1]$, for a 2-dimensional outcome $Y \in \rb^2$. \textbf{Left:} heteroskedastic Gaussian noise. \textbf{Right:} heteroskedastic exponential noise.}
    \label{figure:3D:illustrative}
\end{figure}

Given a feature vector $X \in \mathcal{X}$ and a response $Y \in \mathcal{Y} \subseteq \mathbb{R}^k$, we propose learning two feature-dependent functions: $f(X) \in \mathbb{R}^k$ and $\Sigma(X) \in \mathbb{R}^{k \times k}$, respectively targeting the conditional mean vector and covariance matrix.
This allows us to estimate a local notion of uncertainty that accounts for potential dependencies between outputs, as can be visualized in \Cref{figure:3D:illustrative}.
We employ the local Mahalanobis distance induced by $\Sigma(X)$, defined by $\sqrt{(f(X) - Y)^\top\Sigma^{-1}(X)(f(X) - Y)}$, as a conformal score. The feature-dependent covariance matrix can be learned jointly with the mean or on top of a fixed, user-specified, point predictor.
This makes our method especially suited for constructing data-dependent conformal sets for existing and potentially black-box point predictors. Moreover, the resulting conformal sets can be computed efficiently simply by evaluating the point predictor $f(X)$ and the local covariance estimate $\Sigma(X)$.

Several multivariate conformal methods already produce ellipsoidal \citep{johnstone2021conformal} or covariance-adaptive regions \citep{messoudi2022ellipsoidal}. Our contribution differs in learning a feature-dependent residual covariance model and yielding a closed-form conformity score that remains computationally cheap at inference and extends naturally to missing outputs, partially observed outputs, and output transformations.

Our framework also allows the user to construct valid multivariate conformal sets when some dimensions of the response vector are missing in the calibration data.
By leveraging quantiles of a learned elliptical distribution, we can score the predictions regardless of the number of observed outputs. This enables the definition of a non-conformity score based on the quantile corresponding to the partially observed output. Using this score, we can infer a prediction set for the full label, even when the training or calibration data are incomplete.

Another extension permitted by our framework is to adjust conformal sets when outputs are partially revealed. For example, given a model predicting $(Y^1, Y^2)$, if $Y^1$ is revealed, our method allows us to refine the predictive set for~$Y^2$. The more correlation between~$Y^1$ and~$Y^2$, the more information we accrue and the more we can reduce the size of the updated predictive set, while preserving coverage guarantees. We also demonstrate how to construct a conformal set on a transformation of the output vector. For example, given a model predicting $(Y^1, Y^2, Y^3) \in \rb^3$ the user might be interested in the transformation $(Y^1, Y^1-Y^2)$. Taking into account the local covariance structure results in more informative predictive sets on the combined variables.

\paragraph{Summary of contributions.} We develop a natural extension of standardized residuals to the multivariate setting, using a learned covariance matrix to estimate the local uncertainty of the predictions.
We show that this approach yields conditionally robust conformal sets that can be computed in closed form, resulting in better empirical conditional coverage.
Furthermore, several extensions of standard conformal prediction are made possible.
More precisely, in this paper we:
\begin{itemize}
    \item Propose a method to build predictive sets that are robust to heteroskedasticity in $\P_{Y|X}$, from scratch or on top of an existing point predictor (Section~\ref{sec:gaussian_cp}).
    \item Characterize a set of all distributions for which standardized residuals yield exact conditional coverage asymptotically (Section~\ref{sec:gaussian_cp}).
    \item Construct valid conformal sets when the calibration data have missing output values (Section~\ref{sec:missing:output}).
    \item Refine the conformal set obtained when outputs are partially revealed (Section~\ref{sec:partially:revealed}), and construct valid conformal sets on output transformations~(\Cref{sec:projection:output}).
\end{itemize}

\paragraph{Related work.}
Conformal prediction for multivariate outputs presents the challenge of modeling joint distributions, as simple hyperrectangle-based methods \citep{romano2019conformalized, neeven2018conformal} often fail to capture dependencies between output coordinates.
More flexible approaches attempt to fit complex shapes using multiple residuals. For example, \citet{tumu2024multi} fit a convex set over all residuals, \citet{johnstone2021conformal} use the empirical covariance matrix, and \citet{klein2025multivariate} apply optimal transport.
However, these methods produce the same conformal set for every model output, thus lacking adaptivity to the input.
Some papers introduce feature-adaptive methods \citep{thurin2025optimal, messoudi2022ellipsoidal}, but they require computing the k-nearest neighbors for each feature vector $X_i$, which is computationally expensive both at conformalization and inference time.

Parametric methods that produce feature-dependent conformal sets \citep{braun2025minimum} aim to overcome this challenge, but typically cannot guarantee conditional coverage. Density-based methods can produce adaptive conformal sets \citep{izbicki2022cd, wang2023probabilistic}, but face the challenge of estimating a full conditional density and the scoring functions become intractable in high dimensions. \citet{plassier2024probabilistic} introduced a sampling-based strategy to alleviate this burden, but the conformal sets obtained are random, which introduces undesirable variability for identical inputs and undermines reproducibility.
Conditional variational autoencoders offer another path, but often involve directional quantile regression or cannot provide conformal sets on top of an existing point predictor \citep{feldman2023calibrated, english2025japan}.

Conformal prediction with missing values has received attention in the recent literature. Most existing work focuses on missingness in the input space \citep{zaffran2023conformal, zaffran2024predictive, kong2025fair}. Some methods address noisy or corrupted labels, but require strong assumptions on the data distribution or only handle label corruption \citep{zhou2025conformal, feldman2025conformal} or partially labeled data \citep{javanmardi2023conformal}.

Conformal prediction under partial observability of outputs resembles hierarchical output settings \citep{principato2024conformal, wieslander2020deep}. For transformation of the outputs, it could be approached by applying existing conformal methods directly to transformed outputs, but existing fitted parametric or generative conformal sets do not always provide a cheap post hoc way to transform uncertainty.

\vspace{-1mm}
\paragraph{Background on conformal prediction.} \emph{Split conformal prediction} is a simple and efficient method for constructing marginally valid prediction sets \citep{papadopoulos2002inductive, lei2018distribution, angelopoulos2021gentle}. Given a dataset $\mathcal{D} = \{(X_i, Y_i)\}_{i=1}^n$ sampled from a distribution~$\mathbb{P}_{X, Y}$, with features $X_i \in \mathcal{X}$ and responses $Y_i \in \mathbb{R}^k$, the goal is to build a prediction region $C_\alpha(X_\text{test})$ that satisfies the marginal coverage guarantee
\begin{align}
\label{eq:basic:CP:guarantee}
    \P_{X, Y} \left( Y_\text{test} \in C_\alpha(X_\text{test}) \right) \geq 1 - \alpha \, .
\end{align}
The data are split into a \emph{training set} $\mathcal{D}_1$ with $n_1$ samples and a \emph{calibration set} $\mathcal{D}_2$ with $n_2$ samples. A model $\hat{p}(Y|X)$ is fit on $\mathcal{D}_1$, and $\mathcal{D}_2$ is used to construct prediction sets with required coverage. The key ingredient is a non-conformity score $S(X,Y)\in\rb$, which measures how unusual a response is relative to the model's prediction. Given the empirical $(1-\alpha)$-quantile (with finite-sample correction) of conformal scores
\begin{equation}
\label{eq:q:alpha}
    \widehat{q}_\alpha = \mathrm{Quantile} \Bigg( \frac{\lceil (1 - \alpha)(n_2 + 1) \rceil}{n_2}; \frac{1}{n_2}\sum_{i=1}^{n_2} \delta_{S(X_i, Y_i)} \Bigg),
\end{equation}
we can construct the following region which satisfies the finite-sample guarantee in \eqref{eq:basic:CP:guarantee}:
\begin{align}
\label{eq:definition:sets}
    C_\alpha(X_\text{test}) = \left\{ y \in \mathcal{Y} \mid S(X_\text{test}, y) \leq \widehat{q}_\alpha \right\}.
\end{align}

This coverage guarantee holds on average over the distribution of $X$ and on the calibration data, but not conditionally on a fixed point $X_\text{test}$. This stronger requirement is known as conditional coverage,
\begin{equation}
\label{eq:basic:CP:conditional:guarantee}
    \P_{X, Y} \left( Y_\text{test} \in C_\alpha(X_\text{test}) \mid X_\text{test} \right) \geq 1 - \alpha \, \quad \textrm{almost surely}.
\end{equation}
To improve conditional coverage, the choice of non-conformity score is essential. 
\citet{plassier2025rectifying} show that conditional coverage is achieved when the distribution of $\P_{S(X,Y)|X}$ is independent of $X$. This leaves the challenge of learning scores such that $S(X,Y)\Perp X$.

\paragraph{Notation.}
We denote by $\mathds{1}\{x \in A\}$ the indicator function, equal to $1$ if $x \in A$ and to $0$ if $x \notin A$, for some set $A$. We denote by $S_k^{++}$ the space of $k \times k$ symmetric positive definite matrices\footnote{For simplicity of notation, we assume without loss of generality that the covariance matrix is invertible. In the case of a singular covariance matrix, the density is well defined in a lower-dimensional subspace and can be written using the pseudoinverse and determinant.}. We denote by \(\delta_x\) the Dirac measure at point \(x\), i.e., a probability measure that assigns mass \(1\) to \(x\) and \(0\) elsewhere. We denote by $\mathcal{N}(\cdot \ |\ \mu, \S)$ the density of a normal distribution with mean $\mu$ and covariance $\S$. We use $[k]$ to indicate the discrete set $\{1 , \dots, k \}$. $\|\cdot\|$ denotes the $L_2$-euclidean norm on $\rb^d$, $\|\cdot\|_2$.

\section{Multivariate Standardized Residuals}
\label{sec:gaussian_cp}

\subsection{Method}

Let us consider the setting of multivariate regression: $\{(X_i, Y_i)\}_{i=1}^n$ are $n$ i.i.d.~feature-response pairs sampled from an unknown distribution $\mathbb{P}_{X, Y}$ on $\mathcal{X}\times\mathcal{Y}$ where $\mathcal{Y} \subseteq \mathbb{R}^k$. Given a feature vector $X$, our goal is to generate a predictive set for $Y$ adaptive to the local uncertainty of $\mathbb{P}_{Y \mid X}$. In simpler terms, the size and shape of our conformal set should be adaptive to $X$, capturing the uncertainty on $Y_\text{test}$ that remains given a certain instance $X_\text{test}$.

In the univariate regression setting ($\mathcal{Y} \subseteq \mathbb{R}$), this is commonly achieved using the well-known ``standardized residuals'' non-conformity score \citep{lei2018distribution}:
\[
S_\mathrm{Stand}(X,Y) \coloneq \frac{|Y-f_\theta(X)|}{\sigma_\phi(X)} \, ,
\]
where $f_\theta: \mathcal{X} \to \mathbb{R}$ estimates $Y$ and $\sigma_\phi : \mathcal{X} \to \mathbb{R}_+$ is a learned scalar standard deviation estimate, that allows the size of the conformal set to vary with $X$.  
We generalize this to the multivariate setting. Given a predictor $f_\theta: \mathcal{X} \to \mathbb{R}^k$, we propose normalizing the error residuals using a local variance estimate $\Sigma_\phi(X) \in S_k^{++}$.
We introduce the score $S_{\mathrm{Mah}}$, defined as the Mahalanobis distance between~$Y$ and $f_\theta(X)$ induced by $\Sigma_\phi(X)$:
\[
S_{\mathrm{Mah}}(X,Y) = \| \Sigma_\phi(X)^{-1/2}(Y - f_\theta(X)) \|_2 \,.
\]
This score naturally extends standardized residuals to the multivariate setting $\mathcal{Y} \subseteq \mathbb{R}^k$, in particular we recover $S_{\textrm{Mah}} = S_\textrm{Stand}$ in one dimension. In this work we characterize a set of joint distributions on $(X,Y)$ for which standardized residuals (univariate or multivariate) yield asymptotic conditional coverage.

For a test point $X_\text{test}$, the conformal sets obtained via the Mahalanobis score take the form
\begin{equation*}
    C_\alpha(X_\text{test}) \! = \! \left\{y \in \rb^k, \; S_{\textrm{Mah}}(X_\text{test},y) \leq \hat{q}_\alpha \right\}
    \! = \! \left\{y \in \rb^k, \; \| \Sigma_\phi(X_\text{test})^{-1/2} (y - f_\theta(X_\text{test})) \|_2 \leq \hat{q}_\alpha\right\} ,
\end{equation*} with $\hat{q}_\alpha$ defined in \eqref{eq:q:alpha}.
This set has an intuitive interpretation as an ellipsoid centered at $f_\theta(X_\text{test})$, with a shape parametrized by the local covariance matrix $\Sigma_\phi(X_\text{test})$.
This allows the conformal set to adapt to data heteroskedasticity, in other words, settings where the variance of the residuals depends on $X$, which is commonly observed.
Next, we give theoretical insights for why this score produces conformal sets that approximate conditional coverage.

\subsection{Theoretical guarantees}

In the following, we will consider $(X,Y) \sim \mathbb{P}_{X,Y}$ with $Y \in \mathbb{R}^k$, and assume that $\mathbb{P}_{Y \mid X}$ has a finite second moment $\P_X$-almost surely.
Define the conditional mean $f(X) = \mathbb{E}[Y \mid X]$ and covariance $\Sigma(X) = \mathbb{E}\big[(Y - f(X))(Y - f(X))^\top \mid X \big]$.
Assuming $\Sigma(X)$ is symmetric positive definite $\P_X$-almost surely, we consider the multivariate standardized residual
\[
U = \Sigma(X)^{-1/2}(Y - f(X)) \, .
\]
Simple calculus gives $\mathbb{E}[U \mid X] = 0$ and $\mathbb{E}[U U^\top \mid X] = I_k$ almost surely.
By normalizing $Y-f(X)\mid X$ using its second-order moment, we obtain a random variable $U$ whose first and second conditional moments are independent of $X$. Consequently, using $U$ to define the conformal score is expected to reduce the score's dependence on $X$, thereby improving conditional coverage.
A natural question is therefore whether we can characterize a set of distributions for which normalizing residuals by their local variance yields independence from $X$.
The following proposition provides a sufficient condition for such independence. Proofs for this section are deferred to \Cref{app:missing:proofs}.

\begin{proposition}
\label{prop:W:independant}
    Under the previous assumptions, $U$ is independent of $X$ if and only if there exists a centered random variable $W \in \mathbb{R}^k$ independent of $X$ with covariance $\cov(W)=I_k$ and a function $T(\cdot) \in S_k^{++}$ positive definite almost surely such that $Y=f(X)+T(X)W$.
\end{proposition}
The set of distributions for which $U$ is independent of $X$ covers elliptical distributions \citep{hult2002multivariate}, but also goes beyond those models, as no assumption of ellipticity is required for the noise variable $W$. Furthermore, even for distributions that cannot be written in this form, because the first two moments of $U \mid X$ are independent of $X$, one can expect conditional coverage improvement with our method, as we observe empirically in \Cref{sec:experiments}.

Our next proposition explicitly states when standardized residuals yield conditional coverage.
As in \citet[Theorem 25]{izbicki2022cd}, we require the space $\mathcal{Y}$ to be bounded. This constitutes a relatively mild assumption, given the existence of continuous bijective functions mapping $\mathbb{R}^k$ onto $(-1,1)^k$ (provided that all other assumptions remain valid under these transformations). Although the uniform convergence assumptions are theoretically strong, we empirically demonstrate improvements in conditional coverage in \Cref{sec:experiments}. Finally, \Cref{app:missing:proofs} provides finite-sample results that depend on the estimation quality of the conditional mean and covariance matrix.
\begin{proposition}
\label{prop:asymptotic:convergence}
Considering the setting of \Cref{prop:W:independant}, further assume that the conditional CDF of $\|\S(X)^{-1/2}(Y-f(X))\|^2_2$ is $L$-Lipschitz and strictly increasing $\P_X$-a.s., that $\mathcal{Y}$ is compact, $\inf_{x \in \mathcal{X}} \lambda_{\min}(\Sigma(x)) \geq c > 0$, where $\lambda_{\min}(A)$ is the smallest eigenvalue of a matrix $A$, and that the estimators $\hat{f}$ and $\hat{\Sigma}$ converge uniformly to their true counterparts over $\mathcal{X}$ almost surely as the training set size $n_1 \to \infty$.  The conformal sets obtained through the split conformal framework with the score $S_\textrm{Mah}$ yield almost surely, for all $\alpha\in(0,1)$,
    \[
    \Big|\P(Y\in C_\alpha(X)|X)-(1-\alpha)\Big| \underset{n_1, n_2 \rightarrow + \infty}{\longrightarrow} 0 \, .
    \]
\end{proposition}

\subsection{Learning the local variance}

Now that we motivated the use of the local error covariance $\Sigma_\phi(X)$, the question that remains is how to estimate it?

A simple solution is to approximate $\mathbb{P}_{Y|X}$ with a multivariate elliptical distribution with mean $f_\theta(X)$ and covariance $\Sigma_\phi(X)$.
$f_\theta$ targets the conditional mean $\E[Y \mid X]$, while $\Sigma_\phi$ captures the directional uncertainty in $X$.
The model parameters $\phi$ (if $f_\theta$ is already fixed) or $(\theta,\phi)$ (if the mean and covariance are learned jointly) can be learned via simple maximum likelihood estimation.
For simplicity, we fit a Gaussian model in our experiments $\hat{p}(\cdot|X) = \mathcal{N}(\cdot \mid f_\theta(X), \Sigma_\phi(X))$.
We explain how to parametrize the covariance matrix efficiently and how to fit a low-rank covariance matrix to limit the number of model parameters when the dimension of the output space $k$ is large in \cref{app:density:Gaussian}.


\section{Missing outputs}
\label{sec:missing:output}

In many multivariate regression problems, it is common to encounter missing values in the output vectors used for calibration.
For example, consider a climate modeling application where a system predicts multiple interdependent variables, such as temperature, humidity, and wind speed, at different locations.
Due to sensor malfunctions or data transmission issues, some of these target variables may be missing for certain samples.
Traditional conformal prediction methods require fully observed outputs to construct valid prediction sets, which limits their applicability in such settings.
In this section, we show how our framework can be used to produce valid conformal sets, even with missing values in the calibration data.

\paragraph{Method.}
We revisit the multivariate regression setting: we assume that feature-response pairs $(X, Y)$ are sampled from an unknown distribution $\mathbb{P}_{X, Y}$ over $\mathcal{X}\times\mathcal{Y}$, where $\mathcal{Y} \subseteq \mathbb{R}^k$. Given a feature vector $X_\textrm{test}$, the goal is to construct a predictive set for the outcome $Y_\textrm{test}$.
However, in the calibration set, the response vector $Y_i$ may not be fully observed.
We assume that for each sample, only a random subset of indices $\mathcal{O} \subseteq \{1, \dots, k\}$ is revealed. Formally, we assume that the triples $(X, Y, \mathcal{O})$ are drawn i.i.d. from a joint distribution. For a vector $v \in \mathbb{R}^k$, we denote by $v(\mathcal{O})$ the sub-vector restricted to the indices in $\mathcal{O}$. Thus, the calibration data is formed of tuples $(X_i, \mathcal{O}_i, Y_i(\mathcal{O}_i))$, where $Y(\mathcal{O})$ represents the observed responses.

With our covariance estimation, we can interpret our model as an approximation of the full conditional distribution $\mathbb{P}_{Y|X}$ with a multivariate normal distribution, $\hat{p}(\cdot|X) = \mathcal{N}(\cdot \mid f_\theta(X), \Sigma_\phi(X))$. Notice that this entire section can be generalized to any elliptical distribution.
The model allows us to handle partial observations by marginalizing the Gaussian density to the observed dimensions. Specifically, we define a non-conformity score that dynamically adjusts to the number of observed outputs. 
We describe training strategies to learn the feature-dependent covariance matrix with missing outputs in \cref{app:missing:learning:procedure}.
Once the model is trained, we define the non-conformity score as
\begin{equation*}
    S_\text{Miss}(X, Y, \mathcal{O}) = F^{\chi^2}_{|\mathcal{O}|}\left( \left\| \Sigma_{\phi}(X)_{\mathcal{O}}^{-\frac{1}{2}} \big(Y(\mathcal{O}) - f_{\theta}(X)(\mathcal{O})\big) \right\|_2^2 \right),
\end{equation*}
where $|\mathcal{O}|$ is the number of observed components, $F^{\chi^2}_{r}$ is the CDF of a $\chi^2$ distribution with $r$ degrees of freedom,\footnote{In the general elliptical case, $F^{\chi^2}_{|\mathcal{O}|}$ must be replaced by the cumulative distribution function of the squared Mahalanobis distance for the specific family with dimension $|\mathcal{O}|$.} and $\Sigma_{\phi}(X)_{\mathcal{O}}$ is the submatrix of $\Sigma_{\phi}(X)$ corresponding to the indices in $\mathcal{O}$.
The intuition is as follows: if the model is well-specified and the missingness mechanism is independent of $(X,Y)$, the squared Mahalanobis distance on the observed subspace follows a Chi-squared distribution,
\[
\left\| \Sigma_{\phi}(X)_{\mathcal{O}}^{-\frac{1}{2}} \big(Y(\mathcal{O}) - f_{\theta}(X)(\mathcal{O})\big) \right\|_2^2 \sim \chi^2(|\mathcal{O}|) \, .
\]
By applying the probability integral transform \citep{david1948probability}, we obtain a score that is uniformly distributed, $S_\text{Miss} \sim \mathcal{U}([0,1])$. This transformation projects the error onto a common reference scale, allowing us to compare scores across samples even when the set of observed indices $\mathcal{O}$ varies. In particular, this score is independent of $X$ and of $\mathcal{O}$.

For a test input $X_\text{test}$, we get the prediction region (using the threshold $\hat{q}_\alpha$ defined in \eqref{eq:q:alpha})
\[
    C_{\alpha}(X_\text{test})  =  \Big\{ y(\mathcal{O})  \mid  y  \in  \rb^k, \mathcal{O}  \subseteq  [k], S(X_\textrm{test}, y, \mathcal{O})  \leq  \widehat{q}_\alpha \Big\}.
\]
By construction, this region satisfies the finite-sample coverage guarantee for the observed data:
\[
    \mathbb{P} \large( Y(\mathcal{O}) \in C_{\alpha}( X_\text{test}) \mid \mathcal{D}_{\text{1}} \large) \in \Big[ 1 - \alpha, 1 - \alpha + \frac{1}{n_{\text{2}}+1} \Big).
\]
Finally, to construct a conformal set for the full vector $Y_\textrm{test}$, we simply consider the case where all indices are observed, i.e., $\mathcal{O} = [k]$. Since $Y([k]) = Y$, the full prediction set is
\[
C_{\text{full}, \alpha}(X_\text{test}) = \left\{ y \in \mathbb{R}^k \mid S_\text{Miss}(X_\text{test}, y, [k] \, ) \leq \widehat{q}_\alpha \right\} \, .
\]
This set $C_{\text{full}, \alpha}(X_\textrm{test})$ corresponds to the slice of $C_{\alpha}(X_\textrm{test})$ in $\mathbb{R}^k$.
Calibration is performed on partial observations, so the marginal coverage guarantees hold only for $C_{\alpha}(X_\textrm{test})$. While the marginal coverage guarantee does not necessarily hold for $C_{\text{full}, \alpha}$, empirical results in \Cref{sec:experiments} suggest that the fully observed outcome $Y_\textrm{test}$ is covered with probability close to the desired level $1 - \alpha$.

\begin{remark}
Note that few assumptions on the missingness mechanism are required to guarantee marginal coverage of $C_{\alpha}(X_\textrm{test})$.
It suffices that the missingness mechanism for the test sample coincides with that of the calibration samples, which allows us to go beyond the missing completely at random setting.
\end{remark}

\section{Partially revealed outputs}
\label{sec:partially:revealed}

Consider the problem of predicting the fasting blood glucose~$Y^1$ and the total cholesterol~$Y^2$ for a patient with feature vector~$X$.
Assume that measuring these quantities induces some cost and that one can avoid performing such tests if the confidence intervals are refined enough.
Consider the case where the conformal set obtained after model prediction on $(Y^1, Y^2)$ is not refined enough and we are forced to perform a test for~$Y^1$.
Since the two quantities are often correlated due to underlying physiological relationships, incorporating the information from~$Y^1$ would allow us to refine the uncertainty quantification for~$Y^2$. In a classical predictive setting, integrating the revealed information would require re-training a model for~$Y^2$ using the updated feature vector $Z = (X, Y^1)$, which is computationally expensive and often unrealistic.
Instead, we can leverage the local covariance learned to adapt the conformal set without incurring any additional cost.

\paragraph{Method.}
Assume that we are given a calibration dataset, $\mathcal{D}_2=\{(X_i, Y_i)\}_{i=1}^{n_2}$, of $n_2$ i.i.d. samples drawn from $\P_{X, Y}$. Suppose that for a new test point $X_\text{test}$, we observe partial information about the output $Y_\text{test}$. Specifically, write $Y_\text{test} = (Y_\text{test}^r, Y_\text{test}^h)$ where $Y_\text{test}^r$ is the revealed part and $Y_\text{test}^h$ is the hidden (to be predicted) part.
We wish to construct a set $C_\alpha(X_\text{test}, Y_\text{test}^r)$ such that:
\[
\P\left(Y_\text{test}^h \in C_\alpha(X_\text{test}, Y_\text{test}^r) \right) = 1 - \alpha \, ,
\]
where the probability is over the randomness of the test point $(X_\text{test}, Y_\text{test})$. Leveraging information in $Y_\text{test}^r$ allows us to build a prediction set that adapts to the revealed components.

\begin{figure}[h!]
    \center
    \includegraphics[width=0.24\linewidth]{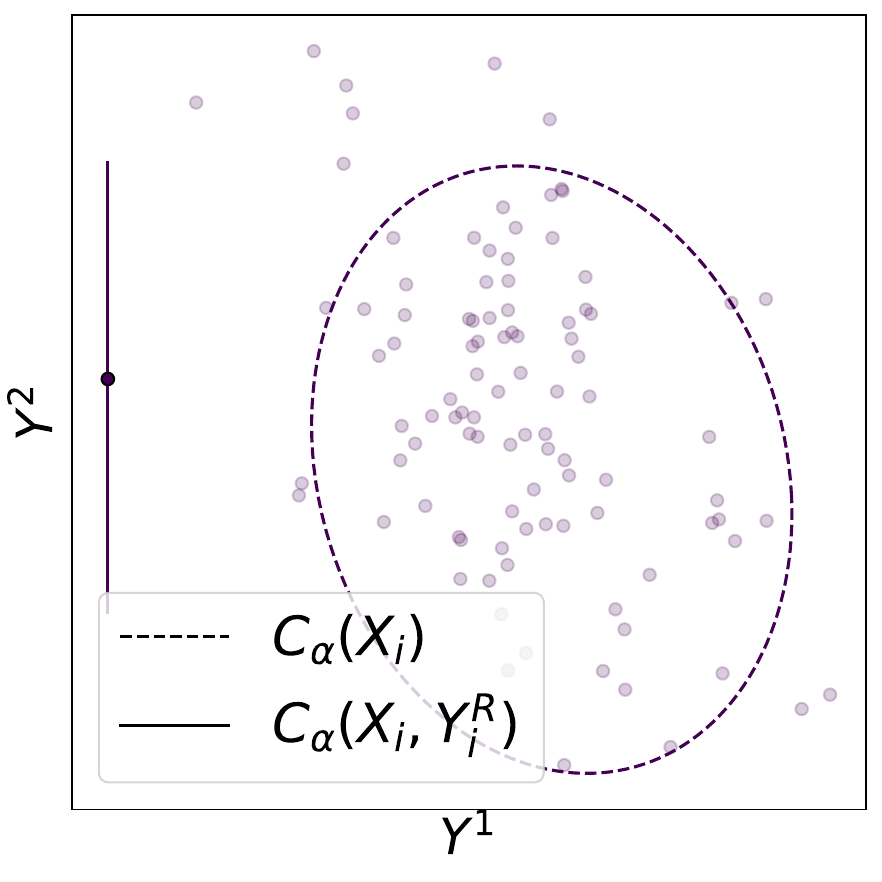}~
\includegraphics[width=0.24\linewidth]{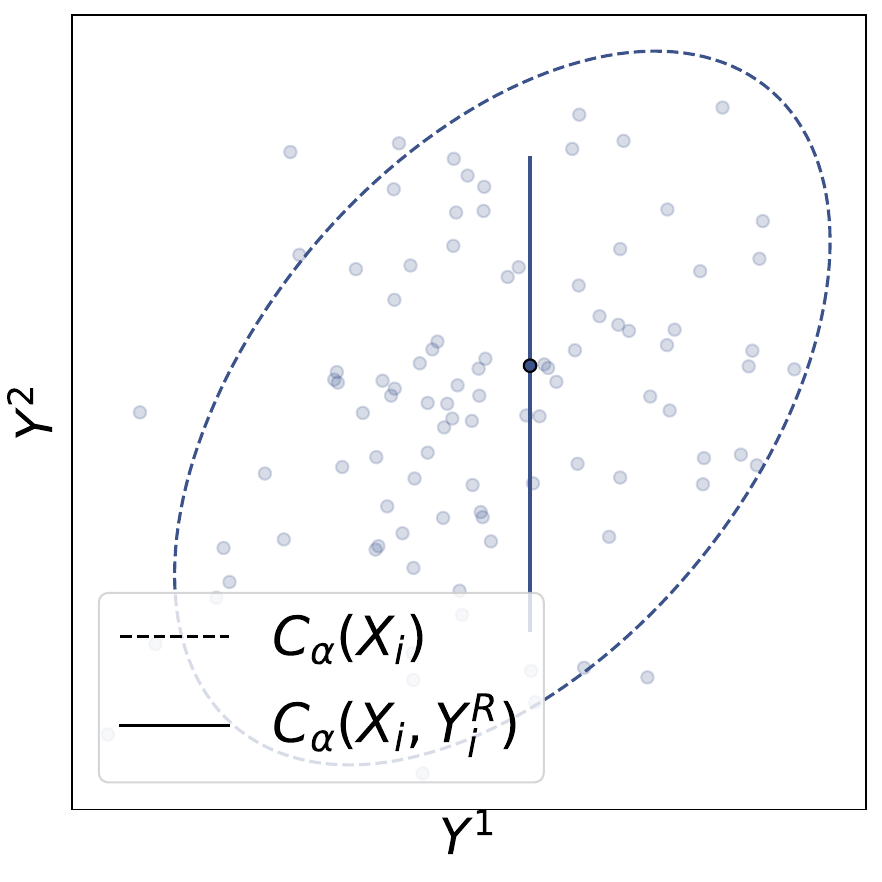}~
\includegraphics[width=0.24\linewidth]{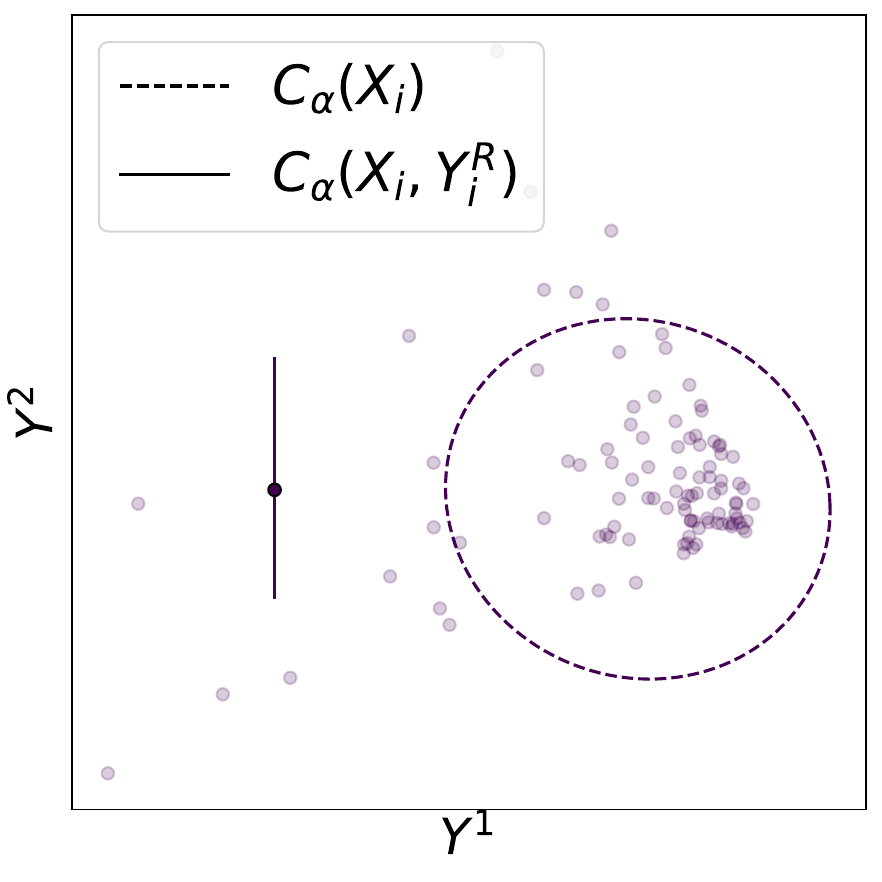}~
\includegraphics[width=0.24\linewidth]{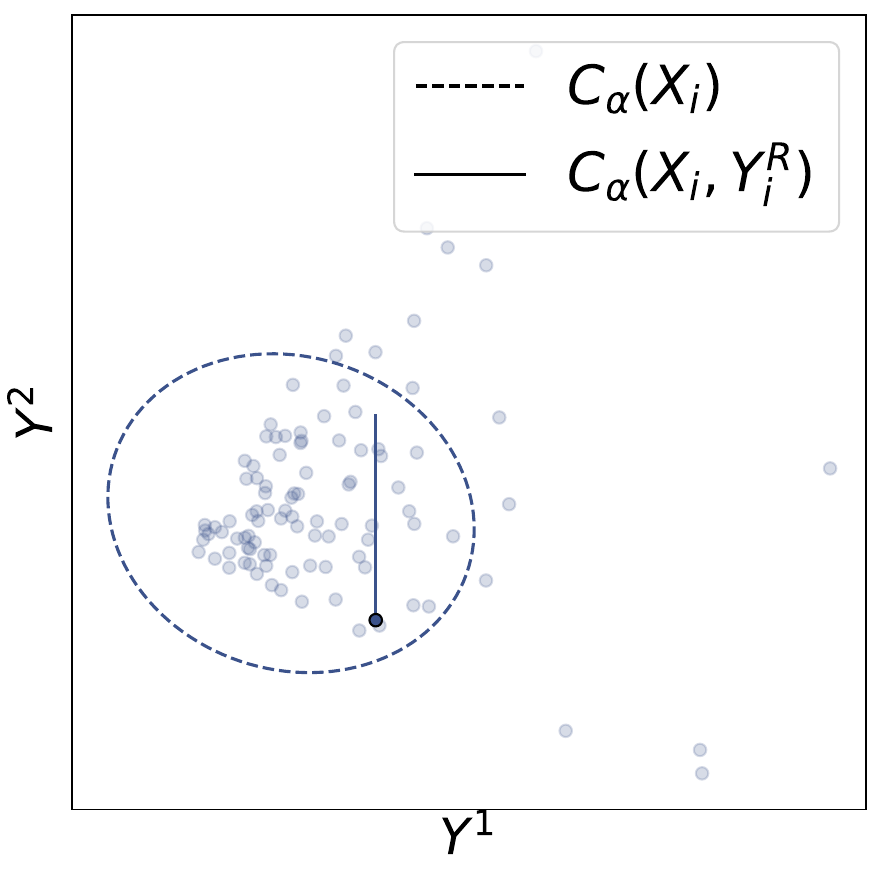}
\vspace{-5mm}
\caption{Conformal sets obtained for partially revealed outputs via the method described in Section~\ref{sec:partially:revealed}.
Data is generated as $Y \sim f(X) + T(X)B$, where $B$ is a standard normal (left) or exponentially distributed (right) random vector and $T(\cdot)$ is a transformation that introduces heteroskedasticity in $X$ (see Appendix~\ref{app:synthetic:datageneration}).
Dark dots show the observed outcome~$Y_i$, while light dots are other samples from $\mathbb{P}_{Y \mid X_i}$ to illustrate the shape of the conditional distribution. 
Each figure is generated for a different $X_i$.
Dashed ellipsoids represent the conformal sets constructed for the full output $(Y^1, Y^2)$ using the Mahalanobis score with coverage parameter $1-\alpha$ set to $90\%$.
Solid-line intervals represent the predictive sets for $Y^2$ obtained with our updated conformal score when the value of $Y^1$ is revealed.}
\label{figure:example:revealed}
\end{figure}

A natural way to incorporate $Y^r$ is through the conditional density $\mathbb{P}_{Y^h \mid X, Y^r}$, computed as
\begin{align*}
    \hat{p}(Y^h|X,Y^r) &= \frac{\hat{p}(Y^h, Y^r|X)}{\hat{p}(Y^r|X)} = \frac{\hat{p}(Y^h, Y^r|X)}{\int \hat{p}(Y^r, y^h|X)\,dy^h} \, .
\end{align*}
Multivariate standardized residuals give us an estimate of the mean and local covariance of residuals which we can use to approximate the conditional density $\P_{Y|X}$ with a Gaussian model $\hat{p}(\cdot|X) = \mathcal{N}(\cdot \mid f(X), \Sigma(X))$.
Plugging this in the equation above, the conditional density of hidden outcomes can be computed in closed form. Denote
\[
\hat{p}(\cdot|X) = \mathcal{N}\left( \, \cdot \, \Big| \,\, \bivec{f^r(X)}{f^h(X)}, \bimat{\Sigma^{rr}(X)}{\Sigma^{rh}(X)}{\Sigma^{hr}(X)}{\Sigma^{hh}(X)} \right),
\]
where we first list the axes of revealed outputs $Y^r$. Conditioning on $Y^r$ preserves Gaussianity. Noting
\[
\begin{cases}
     \tilde{f}(X) &\coloneqq f^h(X) + \Sigma^{hr}(X)\Sigma^{rr}(X)^{-1}(Y^r - f^r(X))\\
     \tilde{\Sigma}(X) &\coloneqq \Sigma^{hh}(X) - \Sigma^{hr}(X)\Sigma^{rr}(X)^{-1}\Sigma^{rh}(X) \, ,
\end{cases}
\]
we have 
\begin{equation*}
\hat{p}(Y^h = \cdot \mid X, Y^r) = \mathcal{N}\left(\cdot \,\middle|\
\tilde{f}(X),\ 
\tilde{\S}(X)
\right).
\end{equation*}
We can apply the same method as in \Cref{sec:gaussian_cp}, using the score
\[
S_\textrm{Revealed}(X, Y^r, Y^h) = \|{\tilde{\Sigma}}(X)^{-1/2}(Y^h - \tilde{f}(X))\|_2 \, .
\]
This yields, with the prediction set $C_\alpha(X_\text{test}, Y^r_\text{test})$ defined in \eqref{eq:definition:sets}, the marginal coverage guarantee
\[
\P \large( Y^h_\text{test} \in C_{\alpha}(X_\text{test}, Y^r_\text{test}) \mid \ \mathcal{D}_1 \large)\\  \in \Big[ 1 - \alpha, 1 - \alpha + \frac{1}{n_2+1} \Big) \, .
\]
We illustrate the effectiveness of this approach in \Cref{figure:example:revealed} with two axes $(Y^1, Y^2)$, when $Y^1$ is revealed.
A naive way to infer sets for $Y^2$ given~$Y^1$ would be to take the section of $C_\alpha(X_\text{test})$ obtained by fixing the value of $Y^1$ to the one revealed.
We see that our method yields more refined predictive sets, which is confirmed by our experiments in \Cref{sec:experiments}.
Notice also that the naive strategy returns empty sets for the purple dots, while our method adapts to the uncertainty in the realization of~$Y^1$.


\section{Projection of the outputs}
\label{sec:projection:output}

In financial portfolio management, suppose we model asset prices $Y^1$ (non-risky), $Y^2$ (gas), and $Y^3$ (electricity). Two strategies yield returns $(R^1, R^2) = (rY^1, pY^2 + qY^3)$, with $r, p, q$ as investment proportions. Since decisions depend on $(R^1, R^2)$ rather than individual $Y^j$, we seek predictive sets for these transformed variables. Constructing separate prediction intervals for each $Y^j$ and propagating them ignores dependencies, leading to overly conservative estimates. Instead, we transfer uncertainty directly to $(R^1, R^2)$ by estimating the density of the transformed output, yielding more reliable predictive sets for decision making.

\begin{figure}[h!]
    \center
    \includegraphics[width=0.24\linewidth]{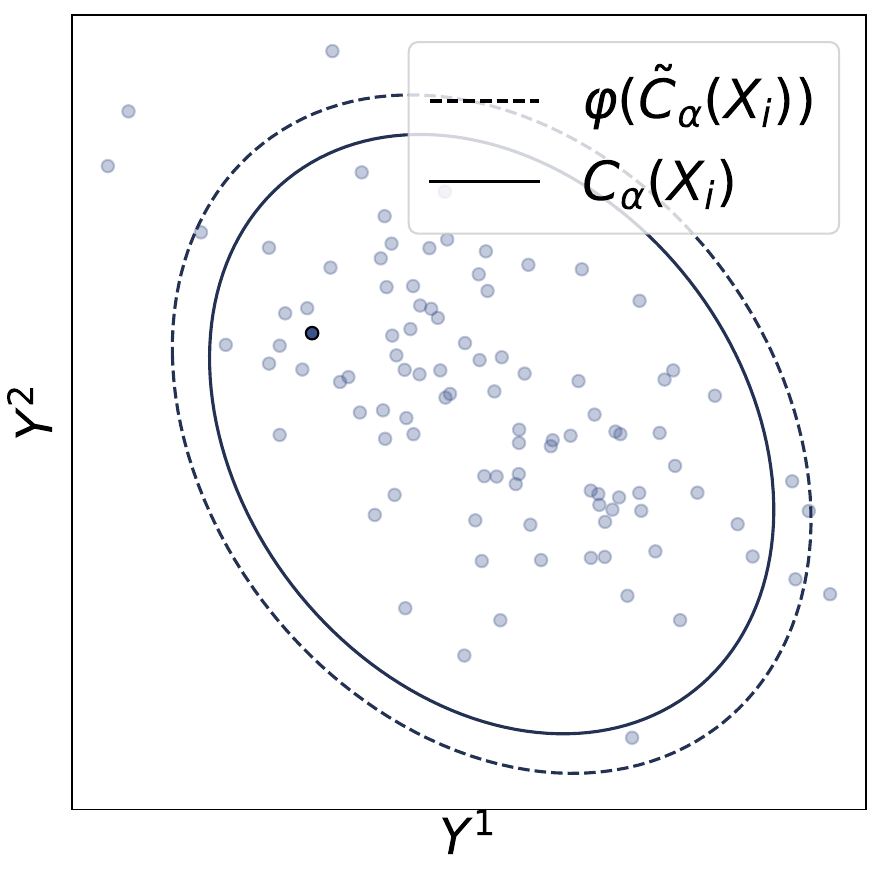}~
\includegraphics[width=0.24\linewidth]{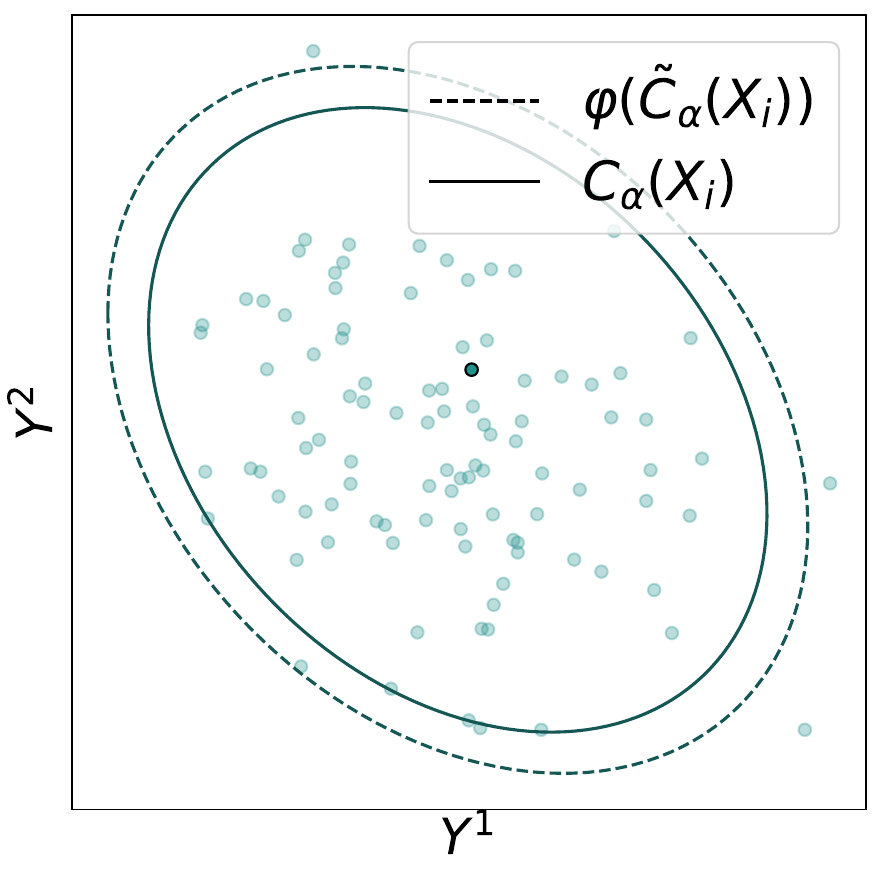}~
\includegraphics[width=0.24\linewidth]{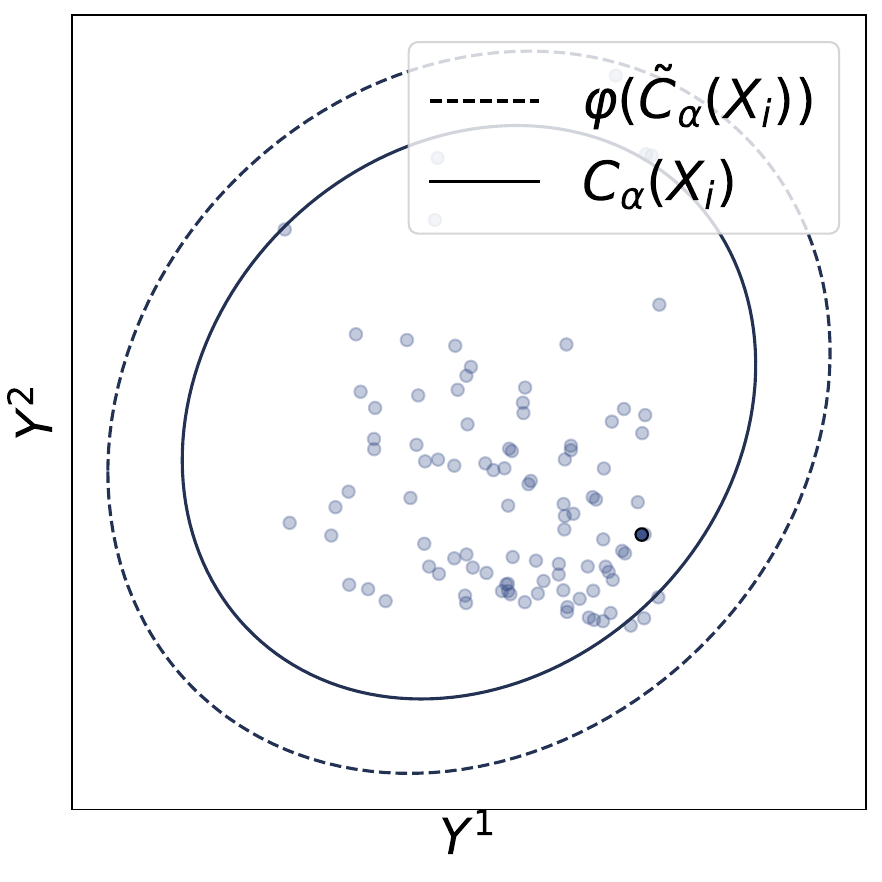}~
\includegraphics[width=0.24\linewidth]{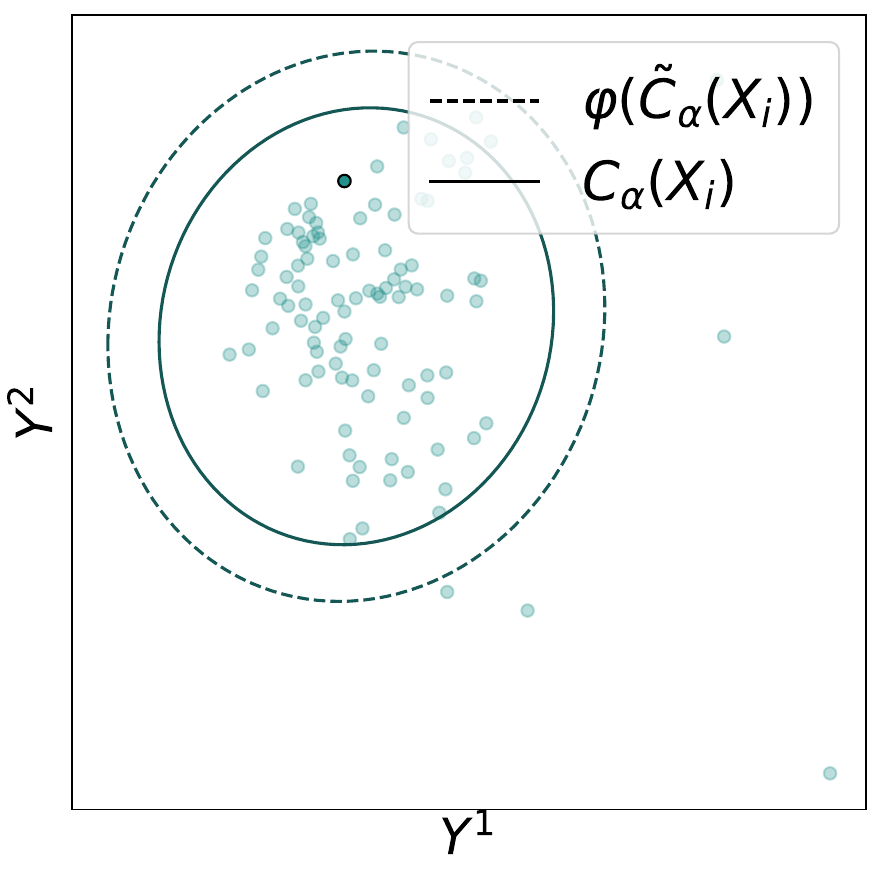} 
    \vspace{-5mm}
    \caption{Comparing conformal sets obtained for projected outputs $\varphi(Y)$, by using the method presented in Section~\ref{sec:projection:output} (full ellipsoids), or by taking the projection $\varphi(\tilde{C}_\alpha(X_i))$ of the Mahalanobis conformal set $\tilde{C}_\alpha(X_i)$ of non-transformed outputs (dashed ellipsoids).
    Data is generated from $Y \sim f(X) + T(X)B$, where $B$ is a standard normal (left)  or exponentially (right) distributed vector and $T(\cdot)$ is a transformation that introduces heteroskedasticity in $X$ (see Appendix~\ref{app:synthetic:datageneration} for details).
    $Y\in \rb^3$, and we consider a projection $\varphi$ of $Y$ in $\rb^2$.
    The coverage parameter $1 - \alpha$ is set to $90\%$.
    We plot the results obtained for four different two different points $X_i$ for each noise.
    Dark dots show the observed transformed label $\varphi(Y_i)$, while light dots depict additional samples from $\varphi(Y) \mid X_i$ to illustrate the shape of the conditional distribution.}
    \label{figure:example:projection}
\end{figure}

\paragraph{Method.}
Assume that we have an elliptical model $\hat{p}(\cdot|X)$ with mean $f_\theta(X)$ and covariance $\S_\phi(X)$ estimating the conditional density $\P_{Y|X}$ and a calibration dataset $\mathcal{D}_2=\{(X_i, Y_i)\}_{i=1}^{n_2}$ of $n_2$ i.i.d. samples drawn from $\P_{X, Y}$.
Given a new feature vector $X_\text{test}$, our goal is to construct a finite-sample valid prediction set for a transformation of the output, denoted $\varphi(\cdot)$. This transformation could be, for example, a projection onto a lower-dimensional subspace, useful when~$Y$ is high-dimensional but only certain components are of interest.
In general, the following inequality holds,
\begin{equation*}
   \P\left(Y_\text{test} \in \tilde{C}_\alpha(X_\text{test})\right) \leq \P\left(\varphi(Y_\text{test}) \in \varphi(\tilde{C}_\alpha(X_\text{test})) \right).
\end{equation*}
So when $\tilde{C}_\alpha(X_\text{test})$ is constructed as in \eqref{eq:definition:sets} the sets $\varphi(\tilde{C}_\alpha(X_\text{test}))$ are conservative (i.e., over-covering) sets for~$\varphi(Y_\text{test})$. If $\varphi$ is an injection, the inequality becomes an equality, and projecting the conformal set directly preserves the coverage level. However, when $\varphi$ is not an injection, the projected set $\varphi(\tilde{C}_\alpha(X))$ may be too large.
To reduce conservativeness, we propose directly modeling the distribution of the transformed variable $\varphi(Y)$ and constructing conformal sets based on its estimated density. This allows us to build tighter sets that still guarantee exact marginal coverage.

Specifically, given our elliptical model $\hat{p}(\cdot|X)$, any linear transformation $\varphi(Y) = MY$ with a deterministic matrix $M \in \mathbb{R}^{p \times k}$ preserves the elliptical structure. The transformed output follows an elliptical distribution parameterized by the mean $Mf_\theta(X)$ and the covariance $M\Sigma_\phi(X)M^\top$.
This allows us to compute level sets for $\varphi(Y)$ using the method described in Section~\ref{sec:gaussian_cp}, leading to the score\footnote{If $M\Sigma_\phi(X)M^\top$ is singular, the distribution of $\varphi(Y)$ is supported on a lower-dimensional linear manifold. In this case, the score is defined by measuring the distance strictly within this subspace, ignoring directions orthogonal to the manifold where the variance is deterministically zero.}
\begin{align*}
    S_\text{LT}(X,Y) = \|( M\Sigma_\phi(X)M^\top)^{-\frac{1}{2}}(MY - Mf_\theta(X)) \|_2 \, .
\end{align*}
With $C_\alpha(X_\text{test})$ the set defined in \eqref{eq:definition:sets}, using the score $S_\text{LT}$ we get the correct marginal coverage
\[
\P_{X, Y} \large( MY \in C_{\alpha}(X) \mid \ \mathcal{D}_1 \large) \in \Big[ 1 - \alpha, 1 - \alpha + \frac{1}{n_2+1} \Big) \, .
\]
In \Cref{app:experiments:baselines}, we compare the sets obtained by transforming the initial conformal sets constructed with the Mahalanobis score $\varphi(\tilde{C}_\alpha(X_\text{test}))$ with the set obtained by conformalizing the transformed density with $S_\text{LT}$, showing that we obtain more refined predictive sets with our method.

\section{Experiments}
\label{sec:experiments}

The code for all our experiments is accessible on github\footnote{\url{https://github.com/ElSacho/Multivariate_Standardized_Residuals}}. 
We present here experiments for empirical conditional coverage on both full and partially revealed outputs and experiments for the missing outputs scheme. We defer experiments for projection of the outputs to \Cref{app:exp:projection}.
We benchmark our method against every baseline targeting conditional coverage in the classical benchmark study of \citet{dheur2025unified} (see Table 1): HPD \citep{izbicki2022cd}, C-PCP and L-CP \citep{dheur2025unified}. Additionally, we consider the PCP method \citep{wang2023probabilistic} and methods designed to generate more interpretable prediction sets from the learned model $f_\theta(\cdot)$, specifically the empirical covariance approach of \cite{johnstone2021conformal}, optimal transport-based sets of \cite{thurin2025optimal}, and MVCS \citep{braun2025minimum}. Details regarding these baselines are in \Cref{app:experiments:baselines}.

\begin{figure}[h]
    \centering
    \subfigure{\includegraphics[width=0.48\columnwidth]{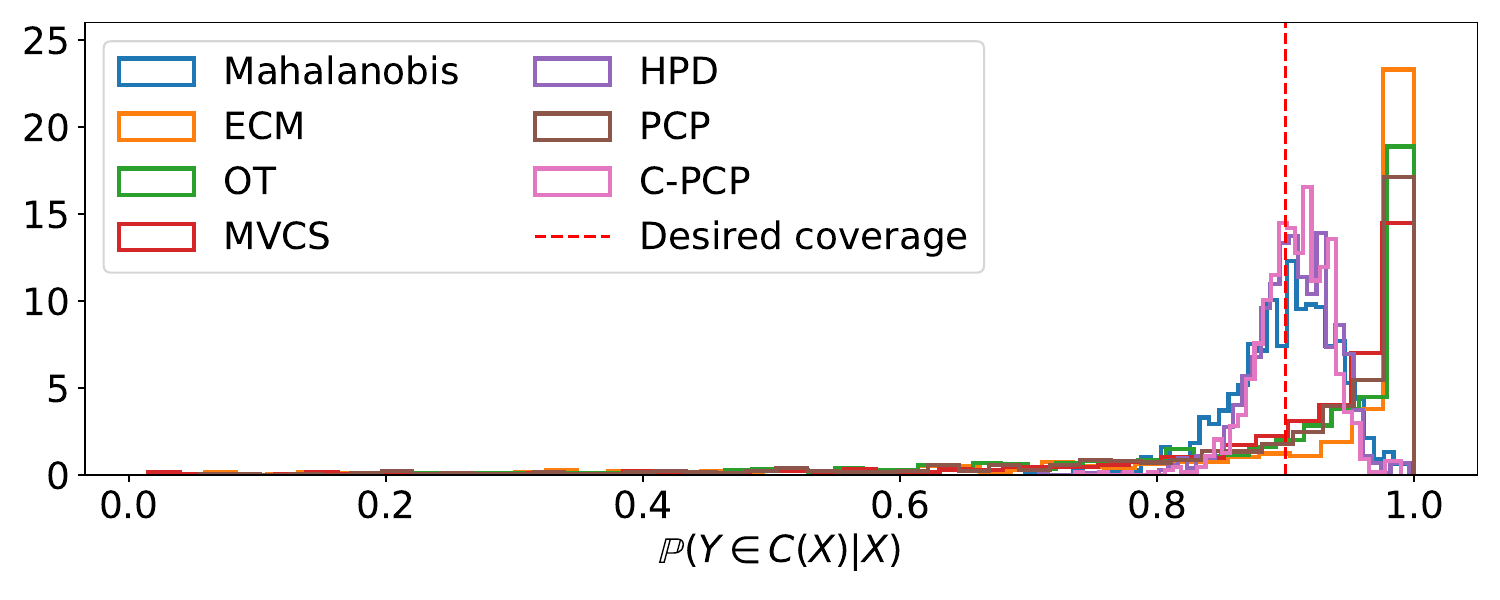}} ~
    \subfigure{\includegraphics[width=0.48\columnwidth]{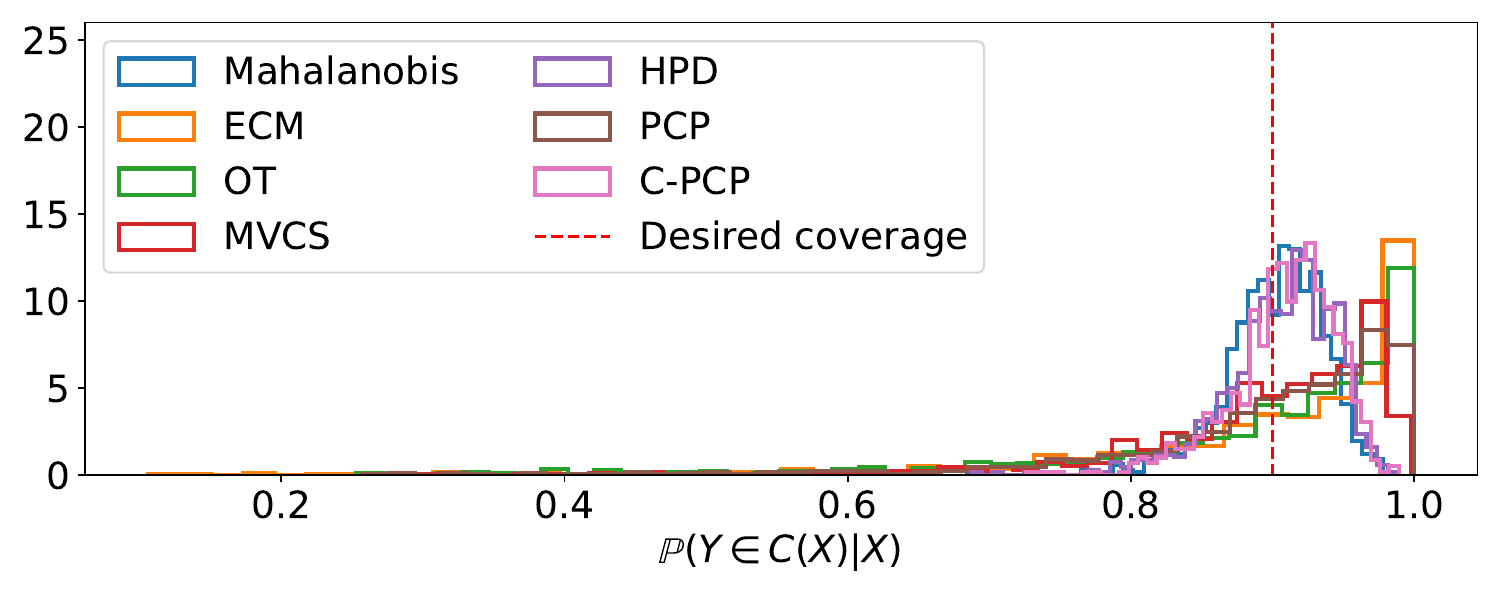}} 
    \vspace{-5mm}
    \caption{Histograms of conditional coverage $\P(Y \in C_\alpha(X_i) \mid X_i)$, with $1-\alpha = 0.90$ (vertical red dashed line) for different conformal prediction methods. Results for 100 test samples per dataset on ten randomly generated synthetic datasets. \textbf{Left:} Heteroskedastic Gaussian noise. \textbf{Right:} Heteroskedastic exponential noise}
    \label{fig:empirical:conditional}
\end{figure}

\paragraph{Synthetic experiments.} We generate synthetic data $Y \sim f(X) + T(X)B$, where $f(\cdot)$ is deterministic, $B$ is a noise vector, $T(\cdot)$ controls heteroskedasticity of the noise, and $Y \in \rb^4$ (see more details in \Cref{app:synthetic:datageneration}). We then train~$f_\theta$ and construct conformal sets ($1-\alpha=0.90$) for 100 randomly sampled test points $X_i$, repeating over 10 datasets. Since $\mathbb{P}_{Y \mid X}$ is known, we estimate $\P(Y \in C_\alpha(X_i)|X_i)$ by sampling $m=1000$ points from $\P_{Y|X_i}$ and compute coverage.
Results, presented in \Cref{fig:empirical:conditional}, show that baseline methods ECM, OT and MVCS (described in \Cref{app:experiments:baselines}) over-cover in some regions to compensate for under-coverage elsewhere, a classical limitation of the ``marginal only'' guarantees of traditional conformal predictions. 

In contrast, just like HPD and other density based methods, standardized residuals achieve approximate conditional coverage around the target level. This holds even when elliptical models are clearly mis-specified, which is consistent with theory.
However, inference time is much quicker with our method, and the produced sets are more interpretable. 

\begin{figure}[h!]
    \center
    \subfigure{\includegraphics[width=0.48\columnwidth]{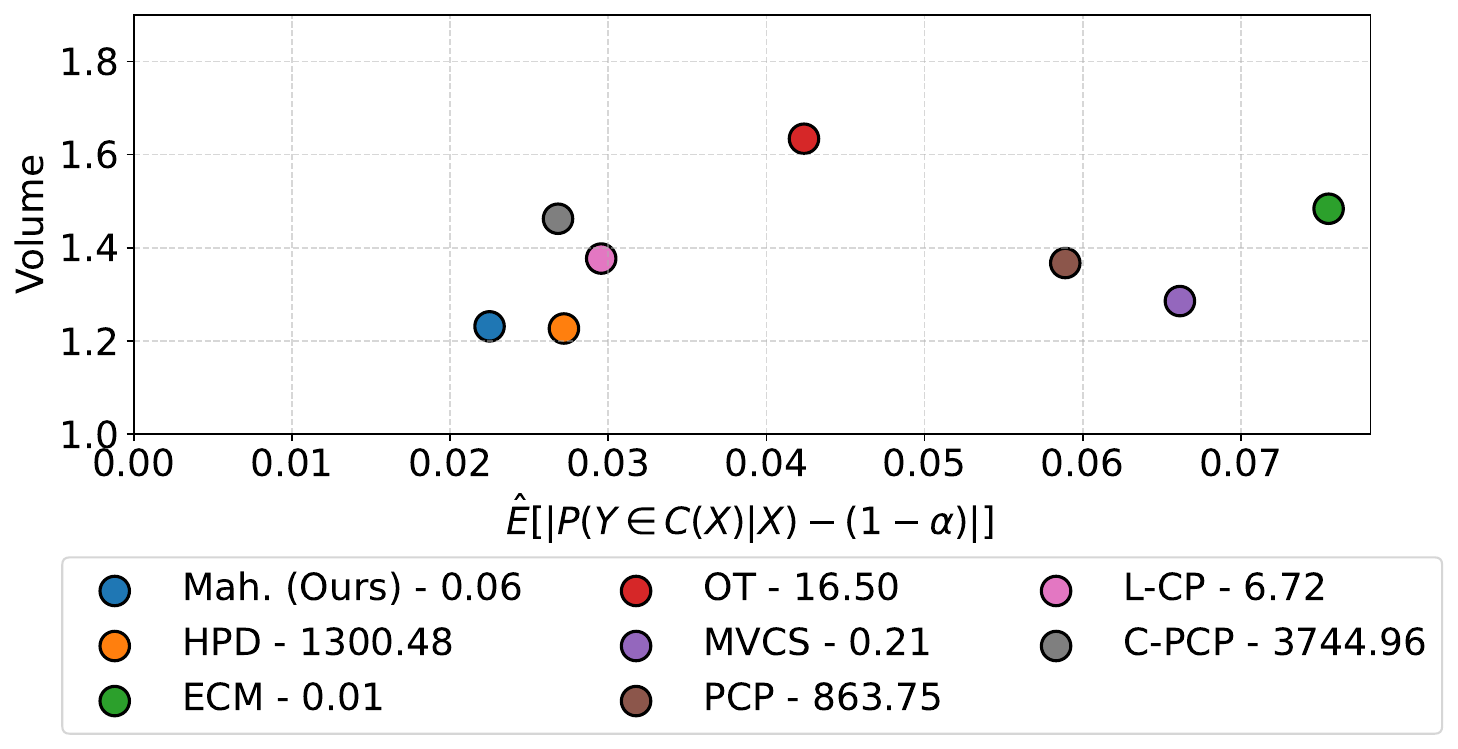}} ~
    \subfigure{\includegraphics[width=0.48\columnwidth]{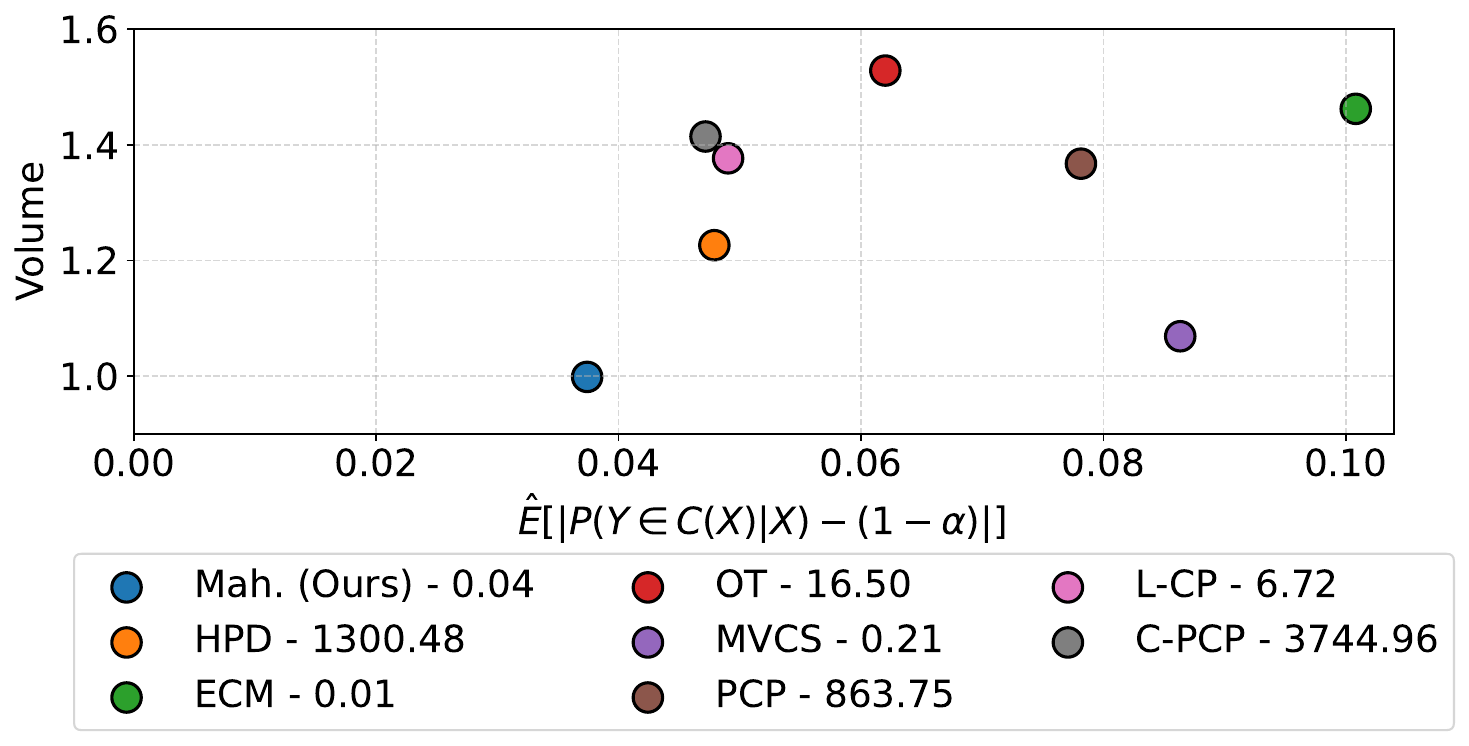}} 
    \vspace{-2mm}
    \caption{Comparison between set size (volume) and conditional coverage deviation. Closer to $(0,0)$ is better.
    Datasets for which baselines (HPD and PCP) failed to produce sets due to large output dimensionality are not considered.
    Numbers displayed next to method names in the legend indicate the compute time to assess coverage normalized per $1,000$ samples.
    \textbf{Left:} Full output for a desired coverage of $90\%$. 
    \textbf{Right:} Revealed output for a desired coverage of $90\%$.}
    \label{figure:size:coverage:tradeoff}
\end{figure}

\paragraph{Conditional coverage.}
\Cref{figure:size:coverage:tradeoff} compares the average set size produced by the different methods with their estimated conditional coverage deviation $\hat \E_X[|\P(Y\in C(X)|X)-(1-\alpha)|]$ computed with the ERT metric \citep{braun2025conditional}.
Results for eight multivariate datasets are averaged and shown as filled dots (details on datasets are provided in \Cref{app:dataset:information}).
Our method generally improves conditional coverage, yielding the smallest estimated average conditional coverage deviation in both the full output from~\Cref{sec:gaussian_cp} and the partially revealed outputs from~\Cref{sec:partially:revealed}. While HPD intervals theoretically produce the tightest sets when the underlying model is perfectly calibrated, the practical difficulty of estimating a full conditional distribution often degrades performance. Because our method requires only the estimation of a covariance matrix, it provides a compelling strategy that is both more robust and less computationally demanding. Furthermore, our strategy is highly computationally efficient; evaluating the coverage for $1,000$ points requires negligible compute time compared to conditional density estimation. Results per dataset are available in Appendices ~\ref{app:exp:full}~and~\ref{app:exp:partially}.
The same conclusion is supported by experiments on output projections in \Cref{app:exp:projection}.

\paragraph{Missing outputs.}
To experiment with missing outputs in the calibration data, we explore two different missingness mechanisms. For the first one, ``missing completely at random'', we randomly add missing values on the output vectors used for calibration, on the same eight datasets.
The number of missing values per output is chosen randomly between $1$ and $k-1$, with $k$ the output dimension.
For the second missing mechanism, we mask the $10\%$ most extreme values of the outputs. 
We report in Table~\ref{tab:coverage:missing:0.9} the marginal coverage achieved by the conformal sets obtained with the method described in Section \ref{sec:missing:output} on a test dataset with the same missing value generating process, but also the coverage obtained for the full test vector with no missing values, for required marginal coverage $1-\alpha = 0.90$  (results for $1-\alpha = 0.95$ are available in Appendix~\ref{app:experiments:missing:outputs}).
For the full test vector, although the constructed conformal sets do not guarantee marginal coverage, empirical results suggest that the achieved coverage remains close to the target level.
Deviations are more pronounced in specific datasets, notably scm1d and scm20d.
Given that this extension fundamentally relies on a Gaussian assumption, we hypothesize that these discrepancies stem from either underlying non-Gaussian distributions or inaccuracies in covariance estimation.
Ultimately, however, the empirical coverage for the full output generally approximates the desired threshold.

\begin{table}[ht]
\centering
\caption{Marginal coverage when conformalizing with missing outputs (desired coverage 0.9).}
\small
\begin{tabular}{lcccc}
\hline
Dataset & \multicolumn{2}{c}{Missing completely at random} & \multicolumn{2}{c}{10\% extreme removed} \\ 
 & With missing & Full output & With missing & Full output \\ \hline
Bias & $90.4 \pm 1.1$ & $85.1 \pm 1.7$ & $90.8 \pm 1.2$ & $90.7 \pm 1.1$ \\ \hline
CASP & $89.8 \pm 0.7$ & $89.3 \pm 0.9$ & $90.2 \pm 0.4$ & $89.7 \pm 0.4$ \\ \hline
House & $89.7 \pm 1.1$ & $87.9 \pm 1.3$ & $89.6 \pm 1.0$ & $87.6 \pm 0.9$ \\ \hline
rf1 & $90.7 \pm 1.2$ & $89.0 \pm 1.6$ & $90.4 \pm 1.3$ & $89.2 \pm 1.3$ \\ \hline
rf2 & $90.5 \pm 1.5$ & $89.1 \pm 1.5$ & $90.1 \pm 1.2$ & $88.7 \pm 1.2$ \\ \hline
scm1d & $89.9 \pm 0.7$ & $81.3 \pm 2.1$ & $89.8 \pm 1.4$ & $86.9 \pm 1.9$ \\ \hline
scm20d & $89.9 \pm 1.2$ & $81.4 \pm 2.2$ & $89.6 \pm 1.4$ & $85.5 \pm 1.3$ \\ \hline
Taxi & $89.9 \pm 0.5$ & $89.3 \pm 0.7$ & $90.0 \pm 0.6$ & $82.7 \pm 0.6$ \\ \hline
\end{tabular}
\label{tab:coverage:missing:0.9}
\end{table}

\section{Conclusions}
\label{sec:conclusion}

In this paper, we introduced \emph{multivariate standardized residuals}, a framework that improves conditional coverage under minimal assumptions. While a limitation of our approach is that strict conditional coverage cannot be theoretically guaranteed for all distributions, we precisely characterize a set of distributions for which it is achieved asymptotically. Our experiments demonstrate that the method achieves robust approximate conditional coverage in practice, constituting an encouraging step toward this overarching goal. Finally, the flexibility of elliptical models allows for several natural extensions of traditional conformal prediction, such as handling missing values in the output vectors, refining conformal sets when the output is partially revealed, and providing valid conformal sets for low-rank transformations of the outputs.

\section*{Acknowledgements}

Authors acknowledge funding from the European Union (ERC-2022-SYG-OCEAN-101071601). Views and opinions expressed are however those of the author(s) only and do not necessarily reflect those of the European Union or the European Research Council Executive Agency. Neither the European Union nor the granting authority can be held responsible for them.

This publication is part of the Chair «Markets and Learning», supported by Air Liquide, BNP PARIBAS ASSET MANAGEMENT Europe, EDF, Orange and SNCF, sponsors of the Inria Foundation.

This work has also received support from the French government, managed by the National Research Agency, under the France 2030 program with the reference «PR[AI]RIE-PSAI» (ANR-23-IACL-0008).


\bibliographystyle{Chicago}
\bibliography{bibliography}

\newpage
\onecolumn
\begin{appendices}
\listofappendices

\counterwithin{figure}{section}
\counterwithin{table}{section}

\crefalias{section}{appendix}
\crefalias{subsection}{appendix}

\newpage
\section{Learning the covariance matrix}
\label{app:density:Gaussian}

Let us consider the setting of multivariate regression: $\{(X_i, Y_i)\}_{i=1}^n$ are $n$ i.i.d. feature-response pairs sampled from an unknown distribution $\P_{X, Y}$ on $\mathcal{X}\times\mathcal{Y}$ where $\mathcal{Y} \subseteq \mathbb{R}^k$. Given a feature vector $X$, our goal is to generate a predictive set for $Y$ using an estimate of the underlying conditional distribution~$\P_{Y|X}$.

A natural idea is to approximate $\P_{Y|X}$ by a multivariate normal distribution,
\[ \hat{p}(\cdot|X) = \mathcal{N}(\cdot \,\,| f_\theta(X), \Sigma_\phi(X)), \]
where $f_\theta: \mathbb{R}^d \to \mathbb{R}^k$ models the mean and $\Sigma_\phi: \mathbb{R}^d \to S_k^{++}$ the covariance. Even if the true distribution is not Gaussian, this model allows us to model the directional uncertainty and effectively learn a feature-dependent covariance matrix. 

A simple way to learn parameters $\theta, \phi$ is to minimize the expected negative log-likelihood:
\[ \mathbb{E}_{\P_{X, Y}}[-\log \hat{p}(Y|X)] \, . \]
We approximate this expectation with the empirical mean on our dataset:
\begin{align}
\label{eq:loss:ce}
    \mathbb{E}_{X, Y}[-\log \hat{p}(Y|X)] 
    &\approx \frac{1}{n} \sum_{i=1}^n -\log \hat{p} (Y_i | X_i) \nonumber \\
    &= \frac{1}{n} \sum_{i=1}^n \log ( (2\pi)^{k/2} \det{\Sigma_\phi(X_i)^{1/2}} ) \\
    &\quad + \frac{1}{2} (Y_i - f_\theta(X_i))^\top \Sigma_\phi(X_i)^{-1} (Y_i - f_\theta(X_i)) \nonumber \\
    &= C^{ste} + \frac{1}{n} \sum_{i=1}^n - \log ( \det \Sigma_\phi(X_i)^{-1/2} ) 
    + \frac{1}{2} \| \Sigma_\phi(X_i)^{-1/2} (Y_i - f_\theta(X_i)) \|_2^2 .
\end{align}

\paragraph{Parameterization.} To ensure that the estimated covariance matrix $\Sigma_\phi(X)$ remains symmetric and positive definite, we parameterize it via its Cholesky decomposition. Specifically, the network outputs the non-zero entries of a lower-triangular matrix $L_\phi(X)$ such that $\Sigma_\phi(X) = L_\phi(X) L_\phi(X)^\top$. To enforce strict positive definiteness, the diagonal elements of $L_\phi(X)$ are passed through a softplus activation shifted by a small $\epsilon$. Proper initialization is critical for training stability; initializing $\Sigma_\phi(X)$ with random weights often results in ill-conditioned matrices and exploding gradients in the first epoch. We therefore initialize the weights of the covariance head such that the off-diagonal terms start near zero and the diagonal terms reflect the marginal standard deviation of the training targets, effectively starting the optimization from a well-behaved independent Gaussian prior.

\begin{remark}[Jointly convex formulation]
    Alternatively, one could employ the canonical parameterization of the exponential family by defining the precision factor $\Lambda_\phi(X) = \Sigma_\phi(X)^{-1/2}$ and the transformed mean parameter~$\eta_\theta(X) = - \Lambda_\phi(X) f_\theta(X)$. With this choice, the negative log-likelihood becomes
    \[
    \mathcal{L} = - \log \det \Lambda_\phi(X) + \frac{1}{2} \| \Lambda_\phi(X) Y + \eta_\theta(X) \|_2^2 + C^{ste} \, ,
    \]
    which is jointly convex in $\eta_\theta$ and $\Lambda_\phi$. In practice, however, given that we parameterize these functions with neural networks, the overall optimization problem remains non-convex with respect to the network weights. We therefore retain the moments parameterization ($f_\theta, \Sigma_\phi$) for clarity, interpretability, and to disentangle the learning of the signal $f_\theta(X)$ from the noise structure $\Sigma_\phi(X)$.
\end{remark}

\begin{remark}[Link with MSE]
Our setup is closely related to learning a point-wise estimate using the mean squared error (MSE), which minimizes:
\[ MSE(f_\theta) = \frac{1}{n} \sum_{i=1}^n \|Y_i - f_\theta(X_i)\|_2^2 \quad . \]
Indeed, taking $\Sigma_\phi \equiv I_k$ (so $\Lambda_\phi \equiv I_k$), minimizing our loss reduces to MSE. By allowing $\Sigma_\phi(\cdot)$ to vary, we enable the model to capture feature-dependent uncertainty while maintaining a point estimate via~$f_\theta$.
\end{remark}

\paragraph{Handling high-dimensional outputs.} In settings where the output dimension $k$ is large, estimating the full $k(k+1)/2$ parameters of the Cholesky factor becomes computationally expensive and prone to overfitting. To mitigate this, we employ a low-rank parameterization inspired by factor analysis: $\Sigma_\phi(X) = D_\phi(X) + V_\phi(X) V_\phi(X)^\top$, where $D_\phi(X) = \text{diag}(d_\phi(X)^2)$ represents feature-dependent independent variances (squared to ensure positivity), and $V_\phi(X) \in \mathbb{R}^{k \times r}$ is a rank-$r$ factor matrix with $r \ll k$. This reduces the parameter complexity from $\mathcal{O}(k^2)$ to $\mathcal{O}(kr)$. Crucially, we avoid explicitly constructing or inverting the full $k \times k$ covariance matrix. Instead, we utilize the Woodbury matrix identity (see e.g., \citealt{horn2012matrix, woodbury1950inverting}) to compute the likelihood and its gradients by inverting only an $r \times r$ matrix, ensuring linear scaling with respect to the output dimension $k$.

\paragraph{Learning a feature-dependent covariance matrix in the presence of missing outputs.} 
\label{app:missing:learning:procedure}
Real-world datasets frequently contain partial observations where entries of $Y_i$ are missing. Under the Gaussian assumption, we can learn from these incomplete samples by maximizing the marginal likelihood of the observed components. For a sample with observed indices $\mathcal{O} \subseteq \{1, \dots, k\}$, we compute the negative log-likelihood using the sub-vector $Y_{\mathcal{O}}$ and the sub-matrix $\Sigma_{\mathcal{O}, \mathcal{O}}$. In our low-rank implementation, this is achieved efficiently without loops or indexing operations by applying a binary mask $M$ to the inverse variance matrix $D^{-1}$. Specifically, we set the inverse variance of missing entries to zero, which mathematically corresponds to infinite uncertainty. This removes the missing dimensions from the precision matrix update while allowing the shared factors $V_\phi(X)$ to learn latent correlations from the available partial data, effectively transferring information between correlated targets even when they are rarely observed together.

An alternative is to impute missing outputs by chosen values. The loss \eqref{eq:loss:ce} is therefore unchanged. The missing outputs can either be replaced by their corresponding predictive values in the vector $f_\theta(X)$, or by an average of all outputs, or more advanced strategies. We refer to \citet{le2021sa} for a more detailed overview of data imputation.

\section{Proofs}
\label{app:missing:proofs}

In the following, we will consider $(X,Y) \sim \mathbb{P}_{X,Y}$ with $Y \in \mathbb{R}^k$, and assume that $\mathbb{P}_{Y \mid X}$ has a finite second moment $\P_X$-almost surely.
Define the conditional mean $f(X) = \mathbb{E}[Y \mid X]$ and covariance $\Sigma(X) = \mathbb{E}\big[(Y - f(X))(Y - f(X))^\top \mid X \big]$.
Assume $\Sigma(X)$ is symmetric positive definite $\P_X$-almost surely.
The norms are either the $L_2$-euclidean norms, for vectors in $\rb^k$, or its induced operator norm for matrices.
Recall that $U = \S(X)^{-1/2}(Y-f(X))$ almost surely. 

\begin{proposition}
\label{app:prop:W:independant}
    Under the previous assumptions, $U$ is independent of $X$ if and only if there exists a centered random variable $W \in \mathbb{R}^k$ independent of $X$ with covariance $\cov(W)=I_k$ and a function $T(\cdot) \in S_k^{++}$ positive definite almost surely such that $Y=f(X)+T(X)W$.
\end{proposition}

\begin{proof}
    Assume that $U \Perp X$. Then $Y=f(X)+\S(X)^{1/2}U$, where $U \Perp X$, $\mathbb{E}[U \mid X] = 0$ and $\mathbb{E}[U U^\top \mid X] = I_k$ almost surely as previously shown. Fixing $T(X)=\S(X)^{-1/2}\in S_k^{++}$ and $W=U$ leads to the first implication.
    
    Assume now that there exists a random variable $W$ independent of $X$ with covariance $\cov(W)=I_k$ and a function $T(X)$ positive definite almost surely such that $Y=f(X)+T(X)W$. Then 
    \begin{align*}
        \cov(Y|X) = \E[(T(X)W)(T(X)W)^\top] = T(X)\cov(W|X)T(X)^T=T(X)T(X)^\top=T(X)^2.
    \end{align*}
    Therefore, $\S(X)^{-1/2}=T(X)^{-1}$ and we get 
    \begin{align*}
        U=\S^{-1/2}(X)\big(f(X)+T(X)W-f(X) \big)=\S^{-1/2}(X)T(X)W = W,
    \end{align*}
    that is independent of $X$.
\end{proof}

\begin{remark}[Sufficiency, but not necessity, for conditional coverage]
The class of distributions for which the method achieves asymptotic conditional coverage is strictly broader than this condition implies. To illustrate this, consider dimension $k = 2$, let $R$ be a continuous positive random variable with $\mathbb{E}[R^2] = 2$, and $X \in \{0, 1\}$ a binary feature. For $X = 0$, let the direction vector $U / \|U\|_2$ be uniformly distributed along the coordinate axes; for $X = 1$, let it be uniformly distributed along the diagonals. Under both conditions, the distributions possess a conditional mean of zero and covariance $I_2$, and the magnitude $\|U\|_2 = R$ has an identical distribution for both values of $X$. Consequently, the Mahalanobis score maintains exact conditional calibration despite $U$ and $X$ not being independent ($U \not\perp X$). Therefore, \Cref{app:prop:W:independant} does not characterize all distributions achieving conditional coverage; rather, it serves as a strictly stronger sufficient condition.
\end{remark}

\begin{lemma}
\label{app:lemma:estimation:error}
    Under the assumptions and characterization of $Y$ of \Cref{prop:W:independant}, further assume that the conditional CDF of $\|\S(X)^{-1/2}(Y-f(X))\|^2$ is $L$-Lipschitz and strictly increasing $\P_X$-a.s., then:
    \begin{align*}
        |\P(Y\in &\, C(X)|X)-(1-\alpha)| \leq L|q(X) - \hat q| \\
        &\quad + C(f, \Sigma, \hat f, \hat \Sigma) \sqrt{\|\hat\Sigma(X) - \Sigma(X)\|+ \|f(X)-\hat f(X)\|  +  \|f(X)-\hat f(X)\|^2},
    \end{align*}
    where the constant $C(f, \Sigma, \hat f, \hat \Sigma)$ is defined as:
    \begin{align*}
        2 \sqrt{2L \big(E[\|Y\!-\!f(X)\|^2|X] \|\hat\Sigma^{-1}(X)\|\|\Sigma^{-1}(X)\|\! + \!2 \|\hat\Sigma^{-1}(X) \| E[ \|Y\!-\!f(X)\| |X] \!+\! \|\hat\Sigma^{-1}(X)\|\big)}.
    \end{align*}
    and $\hat q$ is a constant such that $\P(Y\in C(X)|X) = \P( \|\hat \Sigma^{-1/2}(X)(Y-\hat f(X))\|^2\leq \hat q|X)$, and $q(X)$ is the conditional $1-\alpha$ quantile of $\|\S(X)^{-1/2}(Y-f(X)\|^2$.
\end{lemma}

\begin{proof}
    To ease notation, we suppress the explicit dependence on $(X)$ in the intermediate steps and define the residual $\varepsilon := Y-f(X)$, and the estimation errors $\Delta_f := \hat f(X) - f(X)$ and $\Delta_\Sigma := \hat\Sigma(X) - \Sigma(X)$, for any fixed $X$, $\P_X$-almost surely.

    Let $\hat W = \|\hat \Sigma^{-1/2}(X)(Y-\hat f(X))\|^2$ and $W = \|\Sigma^{-1/2}(X)(Y-f(X))\|^2$. \\
    Because the conditional CDF of $\|\S(X)^{-1/2}(Y-f(X)\|^2$ is $L$-Lipschitz and strictly increasing $\P_X$-a.s., the conditional $1-\alpha$ quantile $q(X)$ of $\|\S(X)^{-1/2}(Y-f(X)\|^2$ is uniquely well defined.
    For $\hat q$ such that $\P(Y\in C(X)|X) = \P(\hat W \leq \hat q|X)$ and any $\delta>0$, we have:
    \begin{align*}
        |\P(Y\in C(X)|X)-(1-\alpha)| &\leq |\P(\hat W \leq \hat q|X) - \P(W \leq \hat q|X)|\\ 
        &\quad + |\P(W \leq \hat q|X) - \P(W \leq q(X)|X)|  \\
        &\leq \P( |W- \hat q| \leq \delta|X) + \P( |W - \hat W| > \delta|X) + L|q(X) - \hat q| \\
        &\leq 2L\delta + \frac{1}{\delta}E[ |W - \hat W| \mid X] + L|q(X) - \hat q|,
    \end{align*}
    where the second inequality follows from the indicator function bound:
    \begin{align*}
        |\P(\hat W \le \hat q) - \P(W \le \hat q)| &= |\E[1_{\{\hat W \le \hat q\}} - 1_{\{W \le \hat q\}}]| \\
        &\leq \E[|1_{\{\hat W \le \hat q\}} - 1_{\{W \le \hat q\}}|] \\
        &\leq \E[1_{\{|W-\hat q|\le \delta\}} + 1_{\{|\hat W - W|\ge \delta\}}].
    \end{align*}

    Next, substituting $Y-\hat f(X) = \varepsilon - \Delta_f$, we can bound the difference $|\hat W - W|$ almost surely:
    \begin{align*}
        |\hat W - W| &= |(\varepsilon - \Delta_f)^\top \hat\Sigma^{-1} (\varepsilon - \Delta_f) - \varepsilon^\top \Sigma^{-1} \varepsilon| \\
        &= |\varepsilon^\top (\hat\Sigma^{-1} - \Sigma^{-1}) \varepsilon - 2\Delta_f^\top \hat\Sigma^{-1} \varepsilon + \Delta_f^\top \hat\Sigma^{-1} \Delta_f| \\
        &\leq \|\varepsilon\|^2 \|\hat\Sigma^{-1} - \Sigma^{-1}\| + 2 \|\Delta_f\| \|\varepsilon\| \|\hat\Sigma^{-1}\| + \|\Delta_f\|^2 \|\hat\Sigma^{-1}\|.
    \end{align*}

    Using the identity $\hat\Sigma^{-1} - \Sigma^{-1} = \Sigma^{-1}(\Sigma-\hat\Sigma)\hat\Sigma^{-1}$, we can bound the inverse covariance difference:
    \begin{align*}
        \|\hat\Sigma^{-1} - \Sigma^{-1}\| \leq \|\Sigma^{-1}\| \|\Delta_\Sigma\| \|\hat\Sigma^{-1}\|.
    \end{align*}

    Taking the conditional expectation and factoring out the common terms yields:
    \begin{align*}
        E[|\hat W - W| \mid X] &\leq E[\|\varepsilon\|^2|X] \|\Delta_\Sigma\|\|\hat\Sigma^{-1}\|\|\Sigma^{-1}\| + 2 \|\hat\Sigma^{-1} \| E[\|\varepsilon\| |X] \|\Delta_f\| + \|\Delta_f\|^2 \|\hat\Sigma^{-1}\| \\
        &\leq K(X) \Big( \|\Delta_\Sigma\| + \|\Delta_f\|  +  \|\Delta_f\|^2 \Big),
    \end{align*}
    where $K(X) := E[\|\varepsilon\|^2|X] \|\hat\Sigma^{-1}\|\|\Sigma^{-1}\| + 2 \|\hat\Sigma^{-1} \| E[ \|\varepsilon\| |X] + \|\hat\Sigma^{-1}\|$.

    Finally, optimizing over $\delta>0$ by setting $\delta = \sqrt{E[|\hat W - W| \mid X] / (2L)}$, we obtain:
    \begin{align*}
        |\P(Y\in C(X)|X)-(1-\alpha)| \leq L|q(X) - \hat q| + 2 \sqrt{2L \cdot K(X)} \sqrt{\|\Delta_\Sigma\| + \|\Delta_f\| + \|\Delta_f\|^2}.
    \end{align*}
    Substituting $K(X)$, $\Delta_\Sigma$, and $\Delta_f$ back into the final expression concludes the proof.
\end{proof}

\begin{proposition}
\label{app:prop:asymptotic:convergence}
    Considering the setting of \Cref{prop:W:independant}, further assume that the conditional CDF of $\|\S(X)^{-1/2}(Y-f(X))\|^2_2$ is $L$-Lipschitz and strictly increasing $\P_X$-a.s., that $\mathcal{Y}$ is compact, $\inf_{x \in \mathcal{X}} \lambda_{\min}(\Sigma(x)) \geq c > 0$, where $\lambda_{\min}(A)$ is the smallest eigenvalue of a matrix $A$, and that the estimators $\hat{f}$ and $\hat{\Sigma}$ converge uniformly to their true counterparts over $\mathcal{X}$ almost surely as the training set size $n_1 \to \infty$.  The conformal sets obtained through the split conformal framework with the score $S_\textrm{Mah}$ yield almost surely, for all $\alpha\in(0,1)$,
    \[
    \Big|\P(Y\in C_\alpha(X)|X)-(1-\alpha)\Big| \underset{n_1, n_2 \rightarrow + \infty}{\longrightarrow} 0 \, .
    \]
\end{proposition}

\begin{proof}

From \Cref{app:lemma:estimation:error}, we get:
    \begin{align*}
        |\P(Y\in& C(X)|X)-(1-\alpha)| \leq L|q(X) - \hat q| \\
        &\quad + C(f, \Sigma, \hat f, \hat \Sigma) \sqrt{\|\hat\Sigma(X) - \Sigma(X)\|+ \|f(X)-\hat f(X)\|  +  \|f(X)-\hat f(X)\|^2},
    \end{align*}

Now, from \Cref{app:prop:W:independant}, $q(X)$ is independent of $X$ and can be considered as a constant parameter~$q^*$. Note that if the assumptions of \Cref{app:prop:W:independant} are not satisfied, we can still expect some conditional coverage improvement because the first two moments of $U$ are independent of $X$. It is reasonable to expect $q(X)$ to be closer to a constant value, but we cannot formally quantify it. 

We are left to prove that $\hat q \to q^*$.

Without loss of generality, we can consider the score $\|\hat \Sigma(X_i)^{-1/2}(Y_i -\hat f(X_i)) \|^2$ instead of its square-root version for the proof, as those two scores are equivalent up to a strictly increasing transformation. Therefore $\ds \hat q \! = \! \mathrm{Quantile} \! \Bigg( \! \frac{\lceil (1 \! - \! \alpha)(n_2 \! + \! 1) \rceil}{n_2}; \frac{1}{n_2}\sum_{i=1}^{n_2} \delta_{\|\hat \Sigma(X_i)^{-1/2}(Y_i -\hat f(X_i)) \|^2} \! \Bigg).$

To prove that $\hat q \to q^*$ almost surely, we explicitly decompose the convergence of the empirical quantile into the convergence of the underlying cumulative distribution functions (CDFs). 

Let $W_i = \|\Sigma(X_i)^{-1/2}(Y_i - f(X_i))\|^2$ be the true Mahalanobis scores, and $\hat W_i = \|\hat \Sigma(X_i)^{-1/2}(Y_i -\hat f(X_i)) \|^2$ be the estimated scores on the calibration set. Let $F(w) = \mathbb{P}(W_i \le w)$ denote the marginal CDF of the true scores. Because W is independent of $X$, the conditional distribution of $W_i$ given $X_i$ is identical to its marginal distribution $F$. 

Let $\hat F_{n_2}(w) = \frac{1}{n_2}\sum_{i=1}^{n_2} \mathbf{1}_{\{\hat W_i \le w\}}$ be the empirical CDF of the estimated scores, and $\tilde F_{n_2}(w) = \frac{1}{n_2}\sum_{i=1}^{n_2} \mathbf{1}_{\{W_i \le w\}}$ be the empirical CDF of the true scores. We analyze the uniform convergence of $\hat F_{n_2}(w)$ to $F(w)$ via the triangle inequality:
\[
|\hat F_{n_2}(w) - F(w)| \leq |\hat F_{n_2}(w) - \tilde F_{n_2}(w)| + |\tilde F_{n_2}(w) - F(w)|.
\]

For the second term on the right-hand side, the Glivenko-Cantelli theorem guarantees that $\sup_w |\tilde F_{n_2}(w) - F(w)| \to 0$ almost surely as $n_2 \to \infty$.

For the first term, we bound the difference in indicator functions. For any arbitrary constant $\epsilon > 0$:
\[
|\mathbf{1}_{\{\hat W_i \le w\}} - \mathbf{1}_{\{W_i \le w\}}| \leq \mathbf{1}_{\{|W_i - w| \le \epsilon\}} + \mathbf{1}_{\{|\hat W_i - W_i| > \epsilon\}}.
\]
Averaging over the $n_2$ calibration points and using triangular inequality yields:
\[
|\hat F_{n_2}(w) - \tilde F_{n_2}(w)| \leq \frac{1}{n_2} \sum_{i=1}^{n_2} \mathbf{1}_{\{|W_i - w| \le \epsilon\}} + \frac{1}{n_2} \sum_{i=1}^{n_2} \mathbf{1}_{\{|\hat W_i - W_i| > \epsilon\}}.
\]

Taking the limit as $n_1, n_2 \to \infty$, the first term on the right-hand side converges to the probability mass in the $\epsilon$-neighborhood: $\sup_w\mathbb{P}(|W_i - w| \le \epsilon) =\sup_w F(w+\epsilon) - F(w-\epsilon)\leq 2L\epsilon$ (since F is assumed $L$-Lipschitz). Thus, taking the limit as $\epsilon \to 0$ drives this bounding probability to $0$. 

Because the domain $\mathcal{Y}$ is compact, $Y$, $f(X)$, and $\Sigma(X)$ are bounded. By the uniform convergence of $\hat{\Sigma}$ and the strict positivity of $\lambda_{\min}(\Sigma(X))$, there almost surely exists a finite sample size after which the eigenvalues of $\hat{\Sigma}(X)$ are also uniformly bounded away from zero. On the cone of positive definite matrices with eigenvalues bounded away from zero, the matrix inverse root mapping $A \mapsto A^{-1/2}$ is uniformly Lipschitz. Because polynomial operations and the squared norm are uniformly Lipschitz on bounded domains, the mapping from the estimators to the score is uniformly continuous. Consequently, the uniform convergence of $\hat{f}$ and $\hat{\Sigma}$ guarantees the uniform convergence of the score difference: $\sup_{x \in \mathcal{X}, y \in \mathcal{Y}} |\hat{W} - W| \to 0$ almost surely as $n_1 \to \infty$.

By the definition of the almost sure limit, there exists a finite integer $N_1$ such that for all $n_1 \geq N_1$, $\sup_{x \in \mathcal{X}, y \in \mathcal{Y}} |\hat{W} - W| < \epsilon$ almost surely. Now, consider any calibration set of size $n_2$ with points $(X_i, Y_i) \in \mathcal{X} \times \mathcal{Y}$. Conditional on a training set where $n_1 \geq N_1$, the maximum possible score estimation error for any point in the domain is strictly less than $\epsilon$. Therefore, for all $i \in [n_2]$, we deterministically have $|\hat{W}_i - W_i| < \epsilon$. This implies the indicator variables $\mathds{1}\{|\hat{W}_i - W_i| > \epsilon\}$ are identically zero for all $n_1 \geq N_1$.

Because the indicator variables evaluate to exactly zero for all $n_1 \geq N_1$, their empirical average over the calibration set is also zero, regardless of $n_2$. Evaluating the joint limit superior, the sequence stabilizes at exactly zero before the limits evaluate to infinity, yielding $\limsup_{n_1, n_2 \to \infty} \frac{1}{n_2} \sum_{i=1}^{n_2} \mathds{1}\{|\hat{W}_i - W_i| > \epsilon\} = 0$ almost surely.

Consequently, $\sup_w |\hat F_{n_2}(w) - \tilde F_{n_2}(w)| \to 0$ almost surely, which establishes that $\hat F_{n_2}(w) \to F(w)$ almost surely for all $w$. 

Because the limiting CDF $F$ is continuous and strictly increasing, the almost sure convergence of the CDFs implies the almost sure convergence of their quantiles. Therefore, the empirical quantile $\hat q = \hat F_{n_2}^{-1}(1-\alpha)$ converges almost surely to the true quantile $q^* = F^{-1}(1-\alpha)$ as $n_1, n_2 \to \infty$.

This demonstrates that $L|q(X) - \hat q| = L|q^* - \hat q| \to 0$ almost surely. Combining this with the convergence of the estimation error terms in the second term of the bound in \Cref{app:lemma:estimation:error}, and because $C(f, \S, \hat f,\hat \S)$ is bounded almost surely, we get that the entire right-hand side of the bound vanishes, yielding:
\[
\Big|\mathbb{P}(Y\in C_\alpha(X)|X)-(1-\alpha)\Big| \underset{n_1, n_2 \rightarrow + \infty}{\longrightarrow} 0
\]
almost surely. This completes the proof.
\end{proof}

\begin{proposition}
\label{app:prop:equivalence:HPD}
    When the conditional estimated density is a radially decreasing elliptical model with mean $f_\theta(\cdot)$  and covariance $\S_\phi(\cdot)$, the conformal sets induced by the HPD-split score are equivalent to those obtained with the Mahalanobis score.
\end{proposition}

\begin{proof}
By definition, if the density $\hat{p}(\cdot|X)$ is elliptical with mean $f_\theta(X)$ and dispersion matrix $\Sigma_\phi(X)$, it can be expressed as:
\begin{equation*}
    \hat{p}(y|X) = k \cdot g\left( \| \Sigma_\phi(X)^{-1/2} (y - f_\theta(X)) \|_2^2 \right),
\end{equation*}
where $k$ is a normalization constant and $g: [0, \infty) \to [0, \infty)$ is a strictly decreasing density generator. We can then rewrite the HPD-split score as:
\begin{align}
    S_\text{HPD}(X, Y) &= \P_{y \sim \hat{p}(\cdot|X)}\left( \; \hat{p}(y|X) \geq \hat{p}(Y|X) \; \right) \nonumber \\
    &= \P_{y \sim \hat{p}(\cdot|X)}\left( g\left( \|\Sigma_\phi(X)^{-1/2} (y - f_\theta(X))\|_2^2 \right) \geq g\left( \| \Sigma_\phi(X)^{-1/2} (Y - f_\theta(X)) \|_2^2 \right) \right) \nonumber \\
    &= \P_{y \sim \hat{p}(\cdot|X)}\left( \|\Sigma_\phi(X)^{-1/2} (y - f_\theta(X))\|_2^2 \leq \| \Sigma_\phi(X)^{-1/2} (Y - f_\theta(X)) \|_2^2 \right) \label{eq:HPD-elliptical-inv} \\
    &= F_{Q}\left( \| \Sigma_\phi(X)^{-1/2} (Y - f_\theta(X)) \|_2^2 \right), \label{eq:HPD-elliptical-cdf}
\end{align}
where \eqref{eq:HPD-elliptical-inv} follows from the fact that $g$ is strictly decreasing, reversing the inequality. In \eqref{eq:HPD-elliptical-cdf}, $F_Q$ denotes the cumulative distribution function (CDF) of the quadratic form $Q = (y - f_\theta(X))^\top \Sigma_\phi(X)^{-1} (y - f_\theta(X))$. 

Because $F_Q$ is a CDF of a continuous distribution, it is strictly increasing on its support. Consequently, the HPD score is a monotonic transformation of the Mahalanobis distance. This implies that the conformal sets are identical. We can therefore define the equivalent non-conformity score:
\[
S_{\textrm{Mah}}(X,Y) = \| \Sigma_\phi(X)^{-1/2}(Y - f_\theta(X)) \|_2 \, .
\]
\end{proof}

\Cref{app:prop:equivalence:HPD} demonstrates that when the true data distribution is radially decreasing elliptical and accurately estimated, our method yields conditional sets of minimal size, inheriting this property from \citet[Theorem 25]{izbicki2022cd}. If the true data distribution is non-elliptical, \Cref{prop:W:independant} and \Cref{prop:asymptotic:convergence} establish that our method still achieves asymptotic conditional coverage, although the resulting sets may be overly conservative in size. However, our empirical evaluations in \Cref{sec:app:experiments} demonstrate that this overconservatism does not occur on the tested datasets. This observation suggests either that the underlying data distributions are approximately elliptical, or that the difficulty of learning a full conditional distribution leads to estimation errors that severely degrade performance in competing methods.

\section{Additional experiments}
\label{sec:app:experiments}

\subsection{Hardware.} We ran the experiments on CPUs (Cascade Lake Intel Xeon 5217 with 8 cores).

\subsection{Baselines}
\label{app:experiments:baselines}
We compare our results with competitive baselines that build conformal sets on top of an existing predictive model $f_\theta(\cdot)$. For all strategies, we learn $f_\theta(\cdot)$ by minimizing the mean squared error on a training dataset, then fix the model and construct predictive sets on top of its predictions. We start by describing comparative baselines.

\begin{itemize}
    \item \textbf{HPD} \citep{izbicki2022cd}: Conformalize the highest predictive density (HPD) by using the non-conformity score $S_\text{HPD}(X, Y) =  \P_{y \sim \hat{p}(\cdot|X)}\left( \; \hat{p}(y|X) \geq \hat{p}(Y|X) \; \right)$. The predictive density is trained by minimizing the negative log-likelihood of a multivariate quantile function forecaster \citep{kan2022multivariate}, which employs input convex neural networks \citep{brandon2017input} to parameterize the conditional distribution as the gradient of a convex potential. We use the implementation of \citet{dheur2025unified}. While our approach can be viewed as a subspace of HPD under elliptical models, the standard HPD framework does not impose a prior on density estimation. As a result, computing the induced conformal score becomes challenging, since it requires sampling from the estimated distribution. Moreover, it is difficult to quantitatively assess the quality of the density estimation. When the estimated density is inaccurate, realized outputs $Y$ are assigned low likelihood, which further complicates the estimation of $S_\textrm{HPD}$. This issue arises in our experiments, particularly for datasets with higher output dimensionality.
    \item \textbf{PCP} \citep{wang2023probabilistic}: Draws \( L \) samples \( \tilde Y^{(l)} \sim \hat p_{Y|x} \) with $\hat p_{Y|x}$ the estimated conditional distribution, and defines conformity as the distance to the nearest sample, \( S_{\text{PCP}}(X,Y) = \min_{l\in[L]} \|Y - \tilde Y^{(l)}\|_2 \); the corresponding region is a union of \( L \) balls centered at the sampled points.
    \item \textbf{L-CP} \citep{dheur2025unified}: Defines conformity in a latent space using an invertible conditional generative model \( \hat Q: \mathcal{Z} \times \mathcal{X} \to \mathcal{Y} \). A latent variable \( Z \sim \mathcal{N}(0, I_d) \) is mapped to the output space via \( \hat Q \), and the conformity score is measured in latent space as
    \[
    S_{\text{L-CP}}(X,Y) = \| \hat Q^{-1}(Y; X) \|.
    \]
    The prediction region is obtained by taking a ball of radius \( \hat q \) around the origin in latent space and mapping it back to the output space. This method avoids grid-based directional quantile regression, improving scalability and computational efficiency, and generalizes distributional conformal prediction to multivariate outputs.

    \item \textbf{C-PCP} \citep{dheur2025unified}: Estimates the conditional CDF of the conformity score \( S(X,Y) \),
\[
S_{\text{CDF}}(x,y) = \P(S_W(X,Y) \le S_W(x,y) \mid X = x),
\]
using a Monte Carlo approximation with $K$ samples
\[
S_{\text{ECDF}}(x,y) = \frac{1}{K} \sum_{k =1}^K \mathds{1}[S_W(x,\hat Y^{(k)}) \le S_W(x,y)],
\quad \hat Y^{(k)} \sim \hat F_{Y|x}.
\]
When \( S(x,y) = S_{\text{PCP}}(x,y) \), this yields
\[
S_{\text{C-PCP}}(x,y) = \frac{1}{K} \sum_{k \in [K]} 
1\!\left\{
\min_{l \in [L]} \|\hat Y^{(k)} - \tilde Y^{(l)}\|_2 \le
\min_{l \in [L]} \|y - \tilde Y^{(l)}\|_2
\right\}.
\]
    
    \item \textbf{Empirical covariance matrix (ECM)} \citep{johnstone2021conformal}: This strategy computes the empirical covariance matrix over the residuals of the training dataset $\hat{\S}$ and then uses the score $S(X, Y) = \|\hat{\S}^{-1/2}(Y-f_\theta(X))\|$.
    \item \textbf{Optimal transport (OT)} \citep{thurin2025optimal}: This strategy generalizes the notion of quantiles in multivariate regression through an optimal transport map between the residuals and uniform samples over the $\rb^k$ unit ball.
    \item \textbf{MVCS} \citep{braun2025minimum}: This strategy generates conformal sets through a parametric function that is trained to minimize the total volume of the conformal sets obtained. Traditionally, the conformal set center $f_\theta(\cdot)$ is allowed to change to further minimize the volume of the conformal set obtained.
\end{itemize}

\subsection{Synthetic dataset generation}
\label{app:synthetic:datageneration}
We generate data similarly to \citet{braun2025minimum}, and recall their model here. The model is $Y \sim f(X) + T(X)B$, where $X \sim \mathcal{N}(0, I_d)$ and $Y \in \mathbb{R}^k$. The noise term $B$ follows a fixed distribution and is modulated by the transformation $T(X)$. The transformation function $T(X)$ is defined as \begin{equation} T(X) = r(X) \exp \left( \sum_{i=1}^{K} w_i(X) \log R_i \right), \end{equation} where $r(X)$ scales the transformation, and the summation interpolates between $K$ randomly generated fixed rotation matrices ${R_i}$ in the Lie algebra via the log-Euclidean framework, ensuring valid rotations.

The interpolation weights $w_i(X)$ are determined based on the proximity of $X$ to $K$ anchor points in $\mathbb{R}^d$, following an inverse squared distance scheme: 
\begin{equation*} 
    \tilde{w}_i(X) = \frac{1}{\|X - X_i\|^4}, \quad w_i(X) = \frac{\tilde{w}_i(X)}{\sum_{j} \tilde{w}_j(X)}. 
\end{equation*} 
This formulation allows the covariance of $Y | X$ to vary smoothly with $X$, generating complex, structured heteroskedastic noise patterns.

The radius function is set to $r(X) = \|X\|/2 + X^\top v + 0.15$, and the function $f(X)$ is defined as \begin{equation*} f(X)= 2 (\sin (X^\top \beta) + \tanh((X \odot X)^\top \beta) + X^\top J_2), \end{equation*} where $v$ and $\beta$ are random fixed parameters in $\mathbb{R}^{d\times k}$, and $J_2$ is a coefficient matrix with all entries set to zero except for $(1,1)$ and $(2,2)$, which are set to $1$.

We generate a dataset of $30,000$ samples from this distribution, which is then split into training, validation, calibration, and test sets.

\subsection{Real datasets}
\label{sec:experiments:revealed:real}

For each dataset-method pair, we provide L1-ERT, WSC, average volumes and marginal coverage obtained. We emphasize again that set-size is not the only metric to consider when comparing conformal prediction methods and that it should be studied jointly with conditional coverage.
In particular, a method achieving conditional coverage with very large sets can indicate that the model fails to learn insightful results, whereas a smaller volume with poor conditional coverage indicates a failure to capture heteroskedasticity.

We conduct experiments setting the required coverage level to $1 - \alpha = 0.90$ and $1 - \alpha = 0.95$. Each dataset is split into four subsets: training, validation, calibration, and test—using a (70\%, 10\%, 10\%, 10\%) split.
To ensure robust evaluation, results are averaged over ten independent runs. To normalize for dimensionality, we report volumes scaled by taking their $1/k$-th power, where $k$ is the output dimension.

Data are pre-processed using a quantile transformation on both the features ($X$) and responses ($Y$) using \texttt{scikit-learn} \citep{pedregosa2011scikit}, ensuring normalization across datasets.

\subsection{Conformal set on the full output vector $Y$}
\label{app:exp:full}

See Tables \ref{tab:app:results_alpha01} and \ref{tab:app:results_alpha005}.
We also compare the calibration time required by each algorithm. Since model training differs across strategies and depends on hyperparameter choices, we focus on the computational cost of the calibration stage, which involves computing the score functions. For most methods, the scores admit closed-form expressions, and the associated computation time is negligible compared to HPD and other density based methods C-PCP and PCP. The only scalable density based method is L-CP. In particular, HPD requires estimating probabilities using the learned probabilistic model via Monte Carlo sampling, which introduces a substantial computational burden even when the number of samples is set to a relatively small value of 100\footnote{Increasing the number of samples led to memory exhaustion.}.

Our method significantly and consistently improves conditional coverage, yielding the lowest ERT values. While less pronounced, the improvements can also be observed for WSC. We note that this metric suffers from the curse of dimensionality and requires a large number of samples for reliable estimation, even when the model is perfectly calibrated (see, e.g., \citealp{braun2025conditional}). Overall, the only baselines with comparable conditional coverage are the density based approaches L-CP, C-PCP and HPD. However, those methods are computationally more expensive than ours and do not always scale well to higher dimensions.

Our method has the additional advantage of simplicity, which facilitates implementation in real-world applications. In the context of conformal prediction, density-based approaches rely on the strong assumption that the full conditional distribution is accurately estimated in order to achieve conditional coverage, a requirement that is often unrealistic in practice. Estimating only the conditional covariance remains a nontrivial task, but is substantially less demanding than full conditional density estimation. This relaxation restricts the class of distributions for which conditional coverage can be achieved; however, because the induced learning problem is simpler, empirical results indicate that our approach performs more reliably on average.

\begin{table}[ht]
\centering
\small
\caption{Comparison of conditional coverage ($\alpha=0.1$) for different conformal methods using the ERT metric (lower is better) and WSC (closer to $0.9$ is better); volume, marginal coverage and the running time to assess coverage per $1,000$ samples. Best values in bold, second best underlined. N/A indicates that the method failed to produce valid results for the corresponding dataset due to poor conditional density estimation leading to numerical issues in high dimensions. Experiments are repeated 10 times, and the index number is the standard error across those 10 experiments.}
\resizebox{\linewidth}{!}{
\begin{tabular}{l l c c c c c c c c }
\hline
{\begin{tabular}{c} Dataset \\ {\small [output dim]} \end{tabular}} & Metric & Mahalanobis & HPD & ECM & OT & MVCS & PCP & L-CP & C-PCP \\
\hline
\multirow{5}{*}{{Bias [2]}} & ERT [\%] & $\underline{1.68_{0.45}}$ & $2.00_{0.58}$ & $3.72_{0.51}$ & $3.69_{0.44}$ & $3.05_{0.40}$ & $\mathbf{1.10_{0.51}}$ & $2.51_{0.59}$ & $1.89_{0.45}$ \\
 & WSC & $0.72_{0.01}$ & $0.72_{0.01}$ & $0.72_{0.01}$ & $\underline{0.73_{0.01}}$ & $0.72_{0.01}$ & $0.73_{0.01}$ & $0.71_{0.01}$ & $\mathbf{0.75_{0.01}}$ \\
 & Volume & $\mathbf{1.07_{0.02}}$ & $1.34_{0.03}$ & $1.11_{0.02}$ & $1.28_{0.04}$ & $\underline{1.09_{0.02}}$ & $1.35_{0.01}$ & $1.35_{0.03}$ & $1.47_{0.02}$ \\
 & Coverage & $0.90_{0.01}$ & $0.90_{0.01}$ & $0.90_{0.01}$ & $0.90_{0.01}$ & $0.90_{0.01}$ & $0.90_{0.00}$ & $0.89_{0.01}$ & $0.92_{0.00}$ \\
 & Time [s] & $0.06_{0.02}$ & $556.89_{78.37}$ & $0.01_{0.00}$ & $19.91_{6.73}$ & $0.12_{0.03}$ & $268.81_{33.41}$ & $2.57_{0.02}$ & $1341.40_{312.88}$ \\
\hline
\multirow{5}{*}{{CASP [2]}} & ERT [\%] & $1.89_{0.25}$ & $\underline{1.52_{0.20}}$ & $4.64_{0.17}$ & $4.32_{0.18}$ & $2.31_{0.20}$ & $2.38_{0.17}$ & $1.69_{0.20}$ & $\mathbf{1.38_{0.17}}$ \\
 & WSC & $0.82_{0.00}$ & $\underline{0.84_{0.00}}$ & $0.80_{0.01}$ & $0.80_{0.01}$ & $0.82_{0.01}$ & $0.81_{0.01}$ & $0.82_{0.00}$ & $\mathbf{0.84_{0.00}}$ \\
 & Volume & $1.37_{0.02}$ & $\underline{1.32_{0.01}}$ & $1.57_{0.01}$ & $1.54_{0.02}$ & $\mathbf{1.30_{0.01}}$ & $1.48_{0.01}$ & $1.42_{0.01}$ & $1.51_{0.01}$ \\
 & Coverage & $0.90_{0.00}$ & $0.90_{0.00}$ & $0.90_{0.00}$ & $0.90_{0.00}$ & $0.90_{0.00}$ & $0.90_{0.00}$ & $0.90_{0.00}$ & $0.90_{0.00}$ \\
 & Time [s] & $0.05_{0.00}$ & $2005.97_{47.81}$ & $0.02_{0.00}$ & $2.52_{0.34}$ & $0.45_{0.07}$ & $1992.11_{254.85}$ & $14.51_{0.63}$ & $7879.45_{99.23}$ \\
\hline
\multirow{5}{*}{{House [2]}} & ERT [\%] & $3.01_{0.23}$ & $2.52_{0.30}$ & $6.54_{0.36}$ & $6.56_{0.33}$ & $6.84_{0.41}$ & $4.03_{0.22}$ & $\underline{2.29_{0.26}}$ & $\mathbf{1.92_{0.30}}$ \\
 & WSC & $0.77_{0.01}$ & $\underline{0.80_{0.00}}$ & $0.72_{0.01}$ & $0.72_{0.01}$ & $0.71_{0.01}$ & $0.77_{0.01}$ & $0.80_{0.01}$ & $\mathbf{0.81_{0.01}}$ \\
 & Volume & $\mathbf{1.00_{0.01}}$ & $1.15_{0.01}$ & $1.22_{0.01}$ & $1.25_{0.02}$ & $\underline{1.15_{0.01}}$ & $1.22_{0.01}$ & $1.27_{0.01}$ & $1.80_{0.51}$ \\
 & Coverage & $0.90_{0.00}$ & $0.90_{0.00}$ & $0.90_{0.00}$ & $0.90_{0.00}$ & $0.90_{0.00}$ & $0.90_{0.00}$ & $0.90_{0.00}$ & $0.91_{0.00}$ \\
 & Time [s] & $0.04_{0.02}$ & $1549.27_{154.06}$ & $0.01_{0.00}$ & $0.48_{0.02}$ & $0.15_{0.01}$ & $874.09_{97.38}$ & $7.25_{0.24}$ & $4462.94_{1045.98}$ \\
\hline
\multirow{5}{*}{{rf1 [8]}} & ERT [\%] & $\mathbf{3.02_{0.76}}$ & $4.70_{0.36}$ & $11.24_{0.37}$ & $\underline{3.30_{0.56}}$ & $9.62_{0.36}$ & $8.72_{0.32}$ & $4.84_{0.53}$ & $3.64_{0.27}$ \\
 & WSC & $\underline{0.73_{0.01}}$ & $\mathbf{0.74_{0.01}}$ & $0.53_{0.02}$ & $0.73_{0.01}$ & $0.55_{0.02}$ & $0.64_{0.02}$ & $0.70_{0.02}$ & $0.72_{0.01}$ \\
 & Volume & $\underline{0.28_{0.00}}$ & $0.46_{0.01}$ & $0.60_{0.01}$ & $0.53_{0.02}$ & $\mathbf{0.28_{0.00}}$ & $0.49_{0.01}$ & $0.48_{0.01}$ & $0.49_{0.00}$ \\
 & Coverage & $0.91_{0.00}$ & $0.91_{0.01}$ & $0.90_{0.00}$ & $0.90_{0.00}$ & $0.90_{0.00}$ & $0.90_{0.01}$ & $0.90_{0.00}$ & $0.90_{0.01}$ \\
 & Time [s] & $0.04_{0.01}$ & $460.26_{5.79}$ & $0.01_{0.00}$ & $18.98_{3.74}$ & $0.14_{0.03}$ & $374.45_{18.00}$ & $2.61_{0.16}$ & $1429.94_{87.31}$ \\
\hline
\multirow{5}{*}{{rf2 [8]}} & ERT [\%] & $\mathbf{3.36_{0.77}}$ & $4.73_{0.31}$ & $9.92_{0.47}$ & $5.84_{0.32}$ & $8.37_{0.35}$ & $7.27_{0.55}$ & $4.92_{0.35}$ & $\underline{3.95_{0.37}}$ \\
 & WSC & $0.73_{0.01}$ & $0.73_{0.01}$ & $0.72_{0.01}$ & $\underline{0.74_{0.01}}$ & $0.71_{0.01}$ & $0.71_{0.01}$ & $0.71_{0.01}$ & $\mathbf{0.74_{0.01}}$ \\
 & Volume & $\mathbf{0.35_{0.02}}$ & $0.44_{0.00}$ & $0.67_{0.03}$ & $0.85_{0.02}$ & $\underline{0.38_{0.02}}$ & $0.44_{0.01}$ & $0.43_{0.01}$ & $0.44_{0.01}$ \\
 & Coverage & $0.90_{0.00}$ & $0.91_{0.01}$ & $0.90_{0.01}$ & $0.91_{0.01}$ & $0.90_{0.01}$ & $0.90_{0.00}$ & $0.89_{0.00}$ & $0.91_{0.00}$ \\
 & Time [s] & $0.03_{0.00}$ & $636.24_{197.26}$ & $0.01_{0.00}$ & $35.04_{5.19}$ & $0.07_{0.00}$ & $331.77_{15.72}$ & $3.04_{0.58}$ & $1415.68_{76.14}$ \\
\hline
\multirow{5}{*}{{scm1d [16]}} & ERT [\%] & $\mathbf{3.01_{0.35}}$ & N/A & $12.33_{0.39}$ & $\underline{3.60_{0.27}}$ & $11.43_{0.22}$ & $12.68_{0.29}$ & $3.64_{0.45}$ & $4.11_{0.27}$ \\
 & WSC & $0.73_{0.01}$ & N/A & $0.57_{0.02}$ & $0.73_{0.02}$ & $0.57_{0.02}$ & $0.53_{0.01}$ & $\underline{0.74_{0.01}}$ & $\mathbf{0.75_{0.01}}$ \\
 & Volume & $\underline{1.14_{0.01}}$ & N/A & $1.51_{0.02}$ & $1.67_{0.03}$ & $1.29_{0.01}$ & $\mathbf{1.03_{0.00}}$ & $1.25_{0.01}$ & $1.19_{0.01}$ \\
 & Coverage & $0.89_{0.01}$ & N/A & $0.90_{0.01}$ & $0.89_{0.01}$ & $0.90_{0.00}$ & $0.89_{0.00}$ & $0.90_{0.01}$ & $0.91_{0.01}$ \\
 & Time [s] & $0.05_{0.00}$ & N/A & $0.01_{0.00}$ & $22.68_{6.77}$ & $0.08_{0.01}$ & $349.84_{19.36}$ & $2.73_{0.18}$ & $1940.33_{420.72}$ \\
\hline
\multirow{5}{*}{{scm20d [16]}} & ERT [\%] & $\mathbf{1.39_{0.49}}$ & N/A & $9.29_{0.40}$ & $\underline{2.91_{0.36}}$ & $8.99_{0.28}$ & $9.66_{0.21}$ & $3.14_{0.43}$ & $3.79_{0.45}$ \\
 & WSC & $0.74_{0.01}$ & N/A & $0.70_{0.01}$ & $\mathbf{0.75_{0.01}}$ & $0.70_{0.01}$ & $0.69_{0.01}$ & $\underline{0.74_{0.01}}$ & $0.73_{0.01}$ \\
 & Volume & $\mathbf{1.29_{0.02}}$ & N/A & $1.75_{0.02}$ & $2.61_{0.08}$ & $1.49_{0.02}$ & $\underline{1.46_{0.01}}$ & $1.74_{0.02}$ & $1.56_{0.01}$ \\
 & Coverage & $0.90_{0.00}$ & N/A & $0.90_{0.00}$ & $0.91_{0.00}$ & $0.90_{0.00}$ & $0.90_{0.00}$ & $0.90_{0.01}$ & $0.90_{0.00}$ \\
 & Time [s] & $0.16_{0.07}$ & N/A & $0.01_{0.00}$ & $27.74_{7.46}$ & $0.08_{0.01}$ & $355.04_{19.24}$ & $2.52_{0.18}$ & $1670.13_{376.98}$ \\
\hline
\multirow{5}{*}{{Taxi [2]}} & ERT [\%] & $\underline{0.64_{0.17}}$ & $0.85_{0.16}$ & $2.77_{0.12}$ & $3.68_{0.17}$ & $2.30_{0.13}$ & $1.27_{0.14}$ & $\mathbf{0.62_{0.20}}$ & $0.77_{0.13}$ \\
 & WSC & $0.84_{0.00}$ & $\underline{0.85_{0.00}}$ & $0.80_{0.00}$ & $0.75_{0.01}$ & $0.82_{0.00}$ & $0.84_{0.00}$ & $0.84_{0.00}$ & $\mathbf{0.85_{0.00}}$ \\
 & Volume & $3.33_{0.01}$ & $\mathbf{2.65_{0.11}}$ & $3.45_{0.01}$ & $3.34_{0.02}$ & $3.32_{0.02}$ & $3.46_{0.28}$ & $\underline{3.08_{0.02}}$ & $3.25_{0.01}$ \\
 & Coverage & $0.90_{0.00}$ & $0.90_{0.00}$ & $0.90_{0.00}$ & $0.90_{0.00}$ & $0.90_{0.00}$ & $0.90_{0.00}$ & $0.90_{0.00}$ & $0.91_{0.00}$ \\
 & Time [s] & $0.06_{0.01}$ & $2594.25_{56.33}$ & $0.02_{0.00}$ & $4.67_{0.69}$ & $0.55_{0.11}$ & $2363.92_{303.34}$ & $18.54_{1.13}$ & $9819.80_{92.52}$ \\
\hline
\end{tabular}
}
\label{tab:app:results_alpha01}
\end{table}

\begin{table}[ht]
\centering
\small
\caption{Comparison of conditional coverage ($\alpha=0.05$) for different conformal methods using the ERT metric (lower is better) and WSC (closer to $0.95$ is better); volume, marginal coverage and the running time to assess coverage per $1,000$ samples. Best values in bold, second best underlined. N/A indicates that the method failed to produce valid results for the corresponding dataset due to poor conditional density estimation leading to numerical issues in high dimensions. Experiments are repeated 10 times, and the index number is the standard error across those 10 experiments.}
\resizebox{\linewidth}{!}{
\begin{tabular}{l l c c c c c c c c }
\hline
{\begin{tabular}{c} Dataset \\ {\small [output dim]} \end{tabular}} & Metric & Mahalanobis & HPD & ECM & OT & MVCS & PCP & L-CP & C-PCP \\
\hline
\multirow{5}{*}{{Bias [2]}} & ERT [\%] & $\underline{1.26_{0.24}}$ & $1.61_{0.23}$ & $2.12_{0.20}$ & $1.82_{0.43}$ & $1.28_{0.25}$ & $\mathbf{0.95_{0.37}}$ & $1.60_{0.29}$ & $1.53_{0.46}$ \\
 & WSC & $0.72_{0.02}$ & $0.76_{0.02}$ & $0.75_{0.01}$ & $\underline{0.76_{0.02}}$ & $0.73_{0.01}$ & $0.76_{0.01}$ & $0.74_{0.01}$ & $\mathbf{0.79_{0.03}}$ \\
 & Volume & $\mathbf{1.29_{0.03}}$ & $1.68_{0.04}$ & $1.36_{0.04}$ & $1.47_{0.04}$ & $\underline{1.31_{0.03}}$ & $1.54_{0.02}$ & $1.66_{0.04}$ & $1.79_{0.06}$ \\
 & Coverage & $0.95_{0.00}$ & $0.96_{0.01}$ & $0.95_{0.00}$ & $0.96_{0.00}$ & $0.95_{0.00}$ & $0.95_{0.00}$ & $0.95_{0.00}$ & $0.96_{0.01}$ \\
 & Time [s] & $0.02_{0.00}$ & $808.27_{150.00}$ & $0.01_{0.00}$ & $17.04_{4.22}$ & $0.14_{0.03}$ & $361.15_{43.23}$ & $3.31_{0.03}$ & $1779.19_{430.56}$ \\
\hline
\multirow{5}{*}{{CASP [2]}} & ERT [\%] & $0.97_{0.14}$ & $0.83_{0.17}$ & $2.44_{0.11}$ & $2.03_{0.12}$ & $0.80_{0.12}$ & $\underline{0.74_{0.13}}$ & $1.00_{0.13}$ & $\mathbf{0.71_{0.12}}$ \\
 & WSC & $0.86_{0.01}$ & $\underline{0.87_{0.00}}$ & $0.85_{0.00}$ & $0.86_{0.00}$ & $0.87_{0.00}$ & $0.87_{0.00}$ & $0.86_{0.00}$ & $\mathbf{0.88_{0.00}}$ \\
 & Volume & $\underline{1.73_{0.03}}$ & $1.74_{0.02}$ & $1.97_{0.02}$ & $2.31_{0.04}$ & $\mathbf{1.63_{0.01}}$ & $1.78_{0.01}$ & $1.85_{0.03}$ & $1.92_{0.03}$ \\
 & Coverage & $0.95_{0.00}$ & $0.95_{0.00}$ & $0.95_{0.00}$ & $0.95_{0.00}$ & $0.95_{0.00}$ & $0.95_{0.00}$ & $0.95_{0.00}$ & $0.95_{0.00}$ \\
 & Time [s] & $0.01_{0.00}$ & $679.71_{76.49}$ & $0.00_{0.00}$ & $0.55_{0.08}$ & $0.09_{0.01}$ & $415.35_{58.90}$ & $3.17_{0.14}$ & $1736.05_{19.29}$ \\
\hline
\multirow{5}{*}{{House [2]}} & ERT [\%] & $1.97_{0.24}$ & $\underline{1.20_{0.22}}$ & $3.68_{0.27}$ & $3.49_{0.17}$ & $3.50_{0.13}$ & $2.27_{0.17}$ & $1.57_{0.21}$ & $\mathbf{1.19_{0.16}}$ \\
 & WSC & $0.81_{0.01}$ & $\mathbf{0.83_{0.01}}$ & $0.75_{0.01}$ & $0.75_{0.01}$ & $0.78_{0.01}$ & $0.80_{0.01}$ & $0.82_{0.01}$ & $\underline{0.83_{0.01}}$ \\
 & Volume & $\mathbf{1.23_{0.02}}$ & $1.46_{0.02}$ & $1.61_{0.02}$ & $1.82_{0.05}$ & $\underline{1.41_{0.04}}$ & $1.42_{0.01}$ & $1.52_{0.01}$ & $1.56_{0.02}$ \\
 & Coverage & $0.95_{0.00}$ & $0.96_{0.00}$ & $0.95_{0.00}$ & $0.95_{0.00}$ & $0.95_{0.00}$ & $0.95_{0.00}$ & $0.95_{0.00}$ & $0.95_{0.00}$ \\
 & Time [s] & $0.01_{0.00}$ & $654.83_{142.51}$ & $0.00_{0.00}$ & $0.21_{0.01}$ & $0.07_{0.00}$ & $411.61_{45.24}$ & $3.35_{0.11}$ & $1580.50_{23.19}$ \\
\hline
\multirow{5}{*}{{rf1 [8]}} & ERT [\%] & $\mathbf{1.65_{0.30}}$ & N/A & $6.54_{0.21}$ & $\underline{1.91_{0.21}}$ & $4.64_{0.27}$ & $4.51_{0.28}$ & $2.53_{0.37}$ & $2.52_{0.36}$ \\
 & WSC & $\mathbf{0.76_{0.01}}$ & N/A & $0.54_{0.02}$ & $\underline{0.76_{0.02}}$ & $0.63_{0.03}$ & $0.66_{0.02}$ & $0.69_{0.03}$ & $0.72_{0.03}$ \\
 & Volume & $\underline{0.32_{0.00}}$ & N/A & $0.89_{0.02}$ & $0.55_{0.02}$ & $\mathbf{0.31_{0.01}}$ & $0.52_{0.00}$ & $0.53_{0.01}$ & $0.52_{0.00}$ \\
 & Coverage & $0.96_{0.00}$ & N/A & $0.95_{0.00}$ & $0.96_{0.00}$ & $0.95_{0.00}$ & $0.95_{0.00}$ & $0.95_{0.00}$ & $0.95_{0.00}$ \\
 & Time [s] & $0.04_{0.00}$ & N/A & $0.01_{0.00}$ & $16.44_{3.72}$ & $0.14_{0.03}$ & $414.11_{23.15}$ & $2.86_{0.17}$ & $1588.91_{96.69}$ \\
\hline
\multirow{5}{*}{{rf2 [8]}} & ERT [\%] & $\mathbf{1.80_{0.47}}$ & N/A & $6.11_{0.26}$ & $3.06_{0.25}$ & $4.61_{0.39}$ & $3.75_{0.22}$ & $\underline{2.08_{0.35}}$ & $2.18_{0.27}$ \\
 & WSC & $0.74_{0.01}$ & N/A & $0.73_{0.01}$ & $\underline{0.76_{0.01}}$ & $0.72_{0.01}$ & $0.73_{0.01}$ & $0.72_{0.01}$ & $\mathbf{0.77_{0.02}}$ \\
 & Volume & $\mathbf{0.40_{0.02}}$ & N/A & $0.95_{0.02}$ & $0.87_{0.02}$ & $\underline{0.42_{0.02}}$ & $0.47_{0.01}$ & $0.48_{0.01}$ & $0.48_{0.00}$ \\
 & Coverage & $0.95_{0.00}$ & N/A & $0.95_{0.00}$ & $0.96_{0.00}$ & $0.95_{0.00}$ & $0.95_{0.00}$ & $0.95_{0.00}$ & $0.96_{0.00}$ \\
 & Time [s] & $0.03_{0.00}$ & N/A & $0.01_{0.00}$ & $35.98_{5.25}$ & $0.08_{0.00}$ & $352.59_{19.86}$ & $3.34_{0.63}$ & $1551.95_{78.70}$ \\
\hline
\multirow{5}{*}{{scm1d [16]}} & ERT [\%] & $\mathbf{1.38_{0.26}}$ & N/A & $6.39_{0.19}$ & $\underline{2.05_{0.43}}$ & $6.18_{0.19}$ & $6.54_{0.18}$ & $2.11_{0.22}$ & $3.88_{0.57}$ \\
 & WSC & $0.74_{0.01}$ & N/A & $0.54_{0.01}$ & $\underline{0.76_{0.03}}$ & $0.54_{0.02}$ & $0.51_{0.02}$ & $0.75_{0.01}$ & $\mathbf{0.93_{0.04}}$ \\
 & Volume & $1.29_{0.01}$ & N/A & $1.96_{0.04}$ & $1.69_{0.04}$ & $1.65_{0.04}$ & $\mathbf{1.13_{0.01}}$ & $1.28_{0.01}$ & $\underline{1.26_{0.01}}$ \\
 & Coverage & $0.94_{0.00}$ & N/A & $0.95_{0.00}$ & $0.95_{0.01}$ & $0.95_{0.00}$ & $0.95_{0.00}$ & $0.95_{0.00}$ & $0.98_{0.01}$ \\
 & Time [s] & $0.05_{0.01}$ & N/A & $0.01_{0.00}$ & $28.33_{7.19}$ & $0.08_{0.00}$ & $361.67_{19.17}$ & $2.78_{0.19}$ & $1720.17_{194.29}$ \\
\hline
\multirow{5}{*}{{scm20d [16]}} & ERT [\%] & $\mathbf{0.92_{0.30}}$ & N/A & $5.97_{0.20}$ & $\underline{1.09_{0.28}}$ & $5.47_{0.22}$ & $5.70_{0.20}$ & $1.72_{0.32}$ & $2.99_{0.21}$ \\
 & WSC & $0.76_{0.01}$ & N/A & $0.68_{0.01}$ & $\underline{0.77_{0.01}}$ & $0.69_{0.01}$ & $0.70_{0.01}$ & $\mathbf{0.78_{0.01}}$ & $0.76_{0.01}$ \\
 & Volume & $\mathbf{1.49_{0.02}}$ & N/A & $2.19_{0.04}$ & $2.64_{0.09}$ & $1.87_{0.03}$ & $\underline{1.60_{0.01}}$ & $1.91_{0.02}$ & $1.73_{0.02}$ \\
 & Coverage & $0.95_{0.00}$ & N/A & $0.95_{0.00}$ & $0.96_{0.00}$ & $0.95_{0.00}$ & $0.95_{0.00}$ & $0.96_{0.00}$ & $0.96_{0.00}$ \\
 & Time [s] & $0.05_{0.01}$ & N/A & $0.01_{0.00}$ & $21.72_{5.12}$ & $0.11_{0.02}$ & $393.40_{23.88}$ & $2.82_{0.20}$ & $1839.43_{405.44}$ \\
\hline
\multirow{5}{*}{{Taxi [2]}} & ERT [\%] & $0.56_{0.14}$ & $0.64_{0.16}$ & $1.80_{0.11}$ & $2.13_{0.15}$ & $1.80_{0.12}$ & $\mathbf{0.40_{0.13}}$ & $\underline{0.53_{0.10}}$ & $0.54_{0.16}$ \\
 & WSC & $0.88_{0.00}$ & $\underline{0.89_{0.00}}$ & $0.85_{0.00}$ & $0.82_{0.01}$ & $0.85_{0.00}$ & $0.88_{0.00}$ & $\mathbf{0.89_{0.00}}$ & $0.89_{0.00}$ \\
 & Volume & $4.07_{0.02}$ & $\mathbf{3.24_{0.06}}$ & $4.19_{0.02}$ & $3.95_{0.02}$ & $4.08_{0.02}$ & $3.71_{0.01}$ & $\underline{3.45_{0.01}}$ & $3.84_{0.03}$ \\
 & Coverage & $0.95_{0.00}$ & $0.95_{0.00}$ & $0.95_{0.00}$ & $0.95_{0.00}$ & $0.95_{0.00}$ & $0.95_{0.00}$ & $0.95_{0.00}$ & $0.95_{0.00}$ \\
 & Time [s] & $0.01_{0.00}$ & $400.28_{10.44}$ & $0.00_{0.00}$ & $0.74_{0.11}$ & $0.09_{0.02}$ & $382.11_{49.91}$ & $3.03_{0.18}$ & $1611.38_{12.59}$ \\
\hline
\end{tabular}
}
\label{tab:app:results_alpha005}
\end{table}

\subsection{Missing outputs}
\label{app:experiments:missing:outputs}

We provide in Table~\ref{tab:coverage:missing:0.95} the marginal coverage achieved by the conformal sets obtained with the method described in Section \ref{sec:missing:output} on a test dataset with the same missing value generating process, and the coverage obtained for the full test vector with no missing values, for required marginal coverage $1-\alpha = 0.95$.

\begin{table}[ht]
\centering
\small
\caption{Marginal coverage when conformalizing with missing outputs (desired coverage 0.95).}
\begin{tabular}{l|cc|cc}
\hline
Dataset & \multicolumn{2}{c|}{Missing at random} & \multicolumn{2}{c}{10\% extreme removed} \\ 
 & With missing & Full output & With missing & Full output \\ \hline
Bias & $95.8 \pm 0.6$ & $92.1 \pm 0.9$ & $95.7 \pm 0.7$ & $96.2 \pm 0.6$ \\ \hline
CASP & $95.0 \pm 0.4$ & $93.7 \pm 0.7$ & $95.1 \pm 0.4$ & $95.2 \pm 0.4$ \\ \hline
House & $94.9 \pm 0.8$ & $93.1 \pm 1.1$ & $94.9 \pm 0.9$ & $93.5 \pm 1.0$ \\ \hline
rf1 & $95.6 \pm 0.8$ & $93.4 \pm 1.1$ & $95.6 \pm 0.9$ & $94.6 \pm 1.1$ \\ \hline
rf2 & $95.1 \pm 0.8$ & $93.1 \pm 1.1$ & $95.2 \pm 1.1$ & $94.0 \pm 1.4$ \\ \hline
scm1d & $94.8 \pm 0.8$ & $89.0 \pm 1.9$ & $94.8 \pm 0.7$ & $93.2 \pm 1.2$ \\ \hline
scm20d & $94.7 \pm 0.7$ & $88.8 \pm 1.7$ & $95.3 \pm 1.5$ & $92.9 \pm 1.3$ \\ \hline
Taxi & $95.0 \pm 0.5$ & $93.8 \pm 0.5$ & $94.9 \pm 0.4$ & $88.9 \pm 0.6$ \\ \hline
\end{tabular}
\label{tab:coverage:missing:0.95}
\end{table}

\subsection{Partially revealed information}
\label{app:exp:partially}
To experiment with the setting of partially revealed information, we reveal a portion of the test output and adjust the predictive set on the remaining dimensions.
Results are displayed in Tables~\ref{tab:app:results:revealed:alpha01}~and~\ref{tab:app:results:revealed:alpha005} for a coverage level of $1-\alpha = 0.9$, and $1-\alpha = 0.95$, respectively.
The Mahalanobis strategy uses the score $S_\textrm{Revealed}$ introduced in \Cref{sec:partially:revealed} based on the estimated density of $Y^h|X,Y^r$.
We see that both the minimal volumes and the most conditional strategy are almost always obtained with our method.

\begin{table}[ht]
\centering
\small
\caption{Comparison of conditional coverage ($\alpha=0.1$) for partially revealed outputs for different conformal methods using the ERT metric (lower is better) and WSC (closer to $0.9$ is better); volume, marginal coverage and the running time to assess coverage per $1,000$ samples. Best values in bold, second best underlined. N/A indicates that the method failed to produce valid results for the corresponding dataset due to poor conditional density estimation leading to numerical issues in high dimensions. Experiments are repeated 10 times, and the index number is the standard error across those 10 experiments.}
\resizebox{\linewidth}{!}{
\begin{tabular}{l l c c c c c c c c }
\hline
{\begin{tabular}{c} Dataset \\ {\small [output dim]} \end{tabular}} & Metric & Mahalanobis & HPD & ECM & OT & MVCS & PCP & L-CP & C-PCP \\
\hline
\multirow{5}{*}{{Bias [2]}} & ERT [\%] & $\underline{1.74_{0.32}}$ & $1.98_{0.53}$ & $4.03_{0.44}$ & $3.76_{0.37}$ & $3.31_{0.33}$ & $\mathbf{1.65_{0.36}}$ & $2.39_{0.54}$ & $2.01_{0.40}$ \\
 & WSC & $0.73_{0.01}$ & $0.72_{0.02}$ & $0.72_{0.01}$ & $0.72_{0.01}$ & $0.73_{0.01}$ & $\underline{0.73_{0.01}}$ & $0.71_{0.02}$ & $\mathbf{0.75_{0.01}}$ \\
 & Volume & $\mathbf{0.89_{0.02}}$ & $1.34_{0.03}$ & $\underline{1.05_{0.03}}$ & $1.15_{0.04}$ & $1.05_{0.03}$ & $1.35_{0.01}$ & $1.35_{0.03}$ & $1.47_{0.02}$ \\
 & Coverage & $0.90_{0.00}$ & $0.90_{0.01}$ & $0.90_{0.01}$ & $0.90_{0.01}$ & $0.90_{0.01}$ & $0.90_{0.00}$ & $0.89_{0.01}$ & $0.92_{0.00}$ \\
 & Time [s] & $\underline{0.02_{0.00}}$ & $718.57_{101.13}$ & $\mathbf{0.01_{0.01}}$ & $25.69_{8.68}$ & $0.16_{0.04}$ & $346.85_{43.11}$ & $3.31_{0.03}$ & $1730.84_{403.72}$ \\
\hline
\multirow{5}{*}{{CASP [2]}} & ERT [\%] & $\mathbf{3.40_{0.35}}$ & $4.76_{0.19}$ & $9.31_{0.18}$ & $7.95_{0.21}$ & $5.37_{0.20}$ & $5.01_{0.21}$ & $6.89_{0.21}$ & $\underline{4.12_{0.10}}$ \\
 & WSC & $\mathbf{0.80_{0.00}}$ & $0.74_{0.01}$ & $0.63_{0.01}$ & $0.71_{0.01}$ & $0.72_{0.01}$ & $0.71_{0.01}$ & $0.72_{0.01}$ & $\underline{0.75_{0.01}}$ \\
 & Volume & $\mathbf{0.79_{0.01}}$ & $1.32_{0.01}$ & $1.20_{0.01}$ & $1.03_{0.20}$ & $\underline{0.81_{0.01}}$ & $1.48_{0.01}$ & $1.42_{0.01}$ & $1.51_{0.01}$ \\
 & Coverage & $0.90_{0.00}$ & $0.90_{0.00}$ & $0.90_{0.00}$ & $0.90_{0.00}$ & $0.90_{0.00}$ & $0.90_{0.00}$ & $0.90_{0.00}$ & $0.90_{0.00}$ \\
 & Time [s] & $\underline{0.01_{0.00}}$ & $438.65_{10.45}$ & $\mathbf{0.00_{0.00}}$ & $0.55_{0.07}$ & $0.10_{0.01}$ & $435.62_{55.73}$ & $3.17_{0.14}$ & $1723.04_{21.70}$ \\
\hline
\multirow{5}{*}{{House [2]}} & ERT [\%] & $\mathbf{2.77_{0.27}}$ & $3.56_{0.17}$ & $7.58_{0.29}$ & $7.09_{0.21}$ & $7.56_{0.36}$ & $5.16_{0.26}$ & $\underline{2.98_{0.26}}$ & $3.08_{0.26}$ \\
 & WSC & $0.79_{0.01}$ & $\underline{0.80_{0.00}}$ & $0.72_{0.01}$ & $0.72_{0.01}$ & $0.72_{0.01}$ & $0.76_{0.01}$ & $0.80_{0.00}$ & $\mathbf{0.80_{0.01}}$ \\
 & Volume & $\mathbf{0.74_{0.01}}$ & $1.15_{0.01}$ & $1.09_{0.02}$ & $1.34_{0.06}$ & $\underline{0.98_{0.02}}$ & $1.22_{0.01}$ & $1.27_{0.01}$ & $248.80_{247.51}$ \\
 & Coverage & $0.90_{0.00}$ & $0.90_{0.00}$ & $0.90_{0.00}$ & $0.90_{0.00}$ & $0.90_{0.00}$ & $0.90_{0.00}$ & $0.90_{0.00}$ & $0.91_{0.00}$ \\
 & Time [s] & $\underline{0.01_{0.00}}$ & $716.92_{71.29}$ & $\mathbf{0.00_{0.00}}$ & $0.22_{0.01}$ & $0.07_{0.00}$ & $404.49_{45.06}$ & $3.35_{0.11}$ & $2065.22_{484.03}$ \\
\hline
\multirow{5}{*}{{rf1 [8]}} & ERT [\%] & $\mathbf{3.37_{0.49}}$ & $4.92_{0.38}$ & $12.15_{0.35}$ & $\underline{4.00_{0.56}}$ & $9.91_{0.39}$ & $9.56_{0.39}$ & $4.90_{0.58}$ & $4.63_{0.32}$ \\
 & WSC & $0.72_{0.01}$ & $\underline{0.73_{0.02}}$ & $0.53_{0.02}$ & $\mathbf{0.73_{0.01}}$ & $0.55_{0.02}$ & $0.63_{0.02}$ & $0.70_{0.02}$ & $0.72_{0.02}$ \\
 & Volume & $\mathbf{0.26_{0.00}}$ & $0.46_{0.01}$ & $0.81_{0.01}$ & $0.55_{0.02}$ & $\underline{0.30_{0.00}}$ & $0.49_{0.01}$ & $0.48_{0.01}$ & $0.49_{0.00}$ \\
 & Coverage & $0.90_{0.00}$ & $0.91_{0.01}$ & $0.90_{0.00}$ & $0.90_{0.00}$ & $0.90_{0.00}$ & $0.90_{0.01}$ & $0.90_{0.00}$ & $0.90_{0.01}$ \\
 & Time [s] & $\underline{0.05_{0.01}}$ & $504.67_{6.35}$ & $\mathbf{0.01_{0.00}}$ & $20.81_{4.10}$ & $0.15_{0.03}$ & $410.58_{19.74}$ & $2.86_{0.17}$ & $1567.92_{95.73}$ \\
\hline
\multirow{5}{*}{{rf2 [8]}} & ERT [\%] & $\underline{4.57_{0.29}}$ & $4.96_{0.27}$ & $11.03_{0.46}$ & $6.70_{0.43}$ & $8.68_{0.30}$ & $7.70_{0.49}$ & $4.93_{0.34}$ & $\mathbf{4.30_{0.36}}$ \\
 & WSC & $0.73_{0.01}$ & $0.74_{0.01}$ & $0.71_{0.01}$ & $\mathbf{0.74_{0.01}}$ & $0.70_{0.01}$ & $0.71_{0.01}$ & $0.71_{0.01}$ & $\underline{0.74_{0.01}}$ \\
 & Volume & $\mathbf{0.33_{0.01}}$ & $0.44_{0.00}$ & $0.90_{0.04}$ & $1.16_{0.06}$ & $0.43_{0.03}$ & $inf_{nan}$ & $\underline{0.43_{0.01}}$ & $0.44_{0.01}$ \\
 & Coverage & $0.90_{0.01}$ & $0.91_{0.01}$ & $0.90_{0.01}$ & $0.91_{0.01}$ & $0.90_{0.01}$ & $0.90_{0.00}$ & $0.89_{0.00}$ & $0.91_{0.00}$ \\
 & Time [s] & $\underline{0.04_{0.00}}$ & $697.63_{216.29}$ & $\mathbf{0.01_{0.00}}$ & $38.42_{5.69}$ & $0.08_{0.00}$ & $363.79_{17.24}$ & $3.34_{0.63}$ & $1552.28_{83.49}$ \\
\hline
\multirow{5}{*}{{scm1d [16]}} & ERT [\%] & $\mathbf{2.99_{0.36}}$ & N/A & $12.38_{0.40}$ & $\underline{3.71_{0.32}}$ & $11.65_{0.19}$ & $13.02_{0.33}$ & $3.73_{0.39}$ & $4.30_{0.38}$ \\
 & WSC & $0.73_{0.01}$ & N/A & $0.58_{0.02}$ & $0.73_{0.02}$ & $0.58_{0.01}$ & $0.54_{0.01}$ & $\underline{0.74_{0.01}}$ & $\mathbf{0.75_{0.01}}$ \\
 & Volume & $\underline{1.12_{0.01}}$ & N/A & $1.58_{0.02}$ & $1.70_{0.03}$ & $1.19_{0.02}$ & $\mathbf{1.03_{0.00}}$ & $1.25_{0.01}$ & $1.19_{0.01}$ \\
 & Coverage & $0.90_{0.00}$ & N/A & $0.90_{0.01}$ & $0.89_{0.01}$ & $0.90_{0.00}$ & $0.89_{0.00}$ & $0.90_{0.01}$ & $0.91_{0.01}$ \\
 & Time [s] & $\underline{0.08_{0.00}}$ & N/A & $\mathbf{0.01_{0.00}}$ & $23.14_{6.91}$ & $0.08_{0.01}$ & $356.98_{19.76}$ & $2.78_{0.19}$ & $1979.93_{429.31}$ \\
\hline
\multirow{5}{*}{{scm20d [16]}} & ERT [\%] & $\mathbf{1.73_{0.35}}$ & N/A & $11.20_{0.38}$ & $4.75_{0.44}$ & $10.30_{0.34}$ & $12.08_{0.29}$ & $\underline{3.76_{0.51}}$ & $5.71_{0.44}$ \\
 & WSC & $\underline{0.74_{0.01}}$ & N/A & $0.66_{0.01}$ & $\mathbf{0.75_{0.01}}$ & $0.67_{0.01}$ & $0.65_{0.01}$ & $0.73_{0.01}$ & $0.70_{0.02}$ \\
 & Volume & $\underline{1.24_{0.02}}$ & N/A & $1.80_{0.02}$ & $2.63_{0.08}$ & $\mathbf{0.82_{0.04}}$ & $1.46_{0.01}$ & $1.74_{0.02}$ & $1.56_{0.01}$ \\
 & Coverage & $0.90_{0.00}$ & N/A & $0.90_{0.00}$ & $0.91_{0.00}$ & $0.90_{0.00}$ & $0.90_{0.00}$ & $0.90_{0.01}$ & $0.90_{0.00}$ \\
 & Time [s] & $\underline{0.09_{0.01}}$ & N/A & $\mathbf{0.01_{0.00}}$ & $30.96_{8.33}$ & $0.09_{0.02}$ & $396.25_{21.47}$ & $2.82_{0.20}$ & $1863.99_{420.73}$ \\
\hline
\multirow{5}{*}{{Taxi [2]}} & ERT [\%] & $9.36_{0.17}$ & $8.57_{0.19}$ & $12.98_{0.15}$ & $11.65_{0.21}$ & $12.29_{0.19}$ & $\underline{8.34_{0.20}}$ & $9.65_{0.17}$ & $\mathbf{7.95_{0.20}}$ \\
 & WSC & $0.62_{0.01}$ & $0.65_{0.00}$ & $0.51_{0.01}$ & $0.58_{0.01}$ & $0.51_{0.01}$ & $\underline{0.66_{0.01}}$ & $0.65_{0.01}$ & $\mathbf{0.69_{0.01}}$ \\
 & Volume & $\mathbf{2.61_{0.01}}$ & $\underline{2.65_{0.11}}$ & $3.25_{0.01}$ & $2.67_{0.33}$ & $2.96_{0.02}$ & $3.46_{0.28}$ & $3.08_{0.02}$ & $3.25_{0.01}$ \\
 & Coverage & $0.90_{0.00}$ & $0.90_{0.00}$ & $0.90_{0.00}$ & $0.90_{0.00}$ & $0.90_{0.00}$ & $0.90_{0.00}$ & $0.90_{0.00}$ & $0.91_{0.00}$ \\
 & Time [s] & $\underline{0.01_{0.00}}$ & $423.34_{9.19}$ & $\mathbf{0.00_{0.00}}$ & $0.76_{0.11}$ & $0.09_{0.02}$ & $385.76_{49.50}$ & $3.03_{0.18}$ & $1602.45_{15.10}$ \\
\hline
\end{tabular}
}
\label{tab:app:results:revealed:alpha01}
\end{table}

\begin{table}[ht]
\centering
\small
\caption{Comparison of conditional coverage ($\alpha=0.05$) for partially revealed outputs for different conformal methods using the ERT metric (lower is better) and WSC (closer to $0.95$ is better); volume, marginal coverage and the running time to assess coverage per $1,000$ samples. Best values in bold, second best underlined. N/A indicates that the method failed to produce valid results for the corresponding dataset due to poor conditional density estimation leading to numerical issues in high dimensions. Experiments are repeated 10 times, and the index number is the standard error across those 10 experiments.}
\resizebox{\linewidth}{!}{
\begin{tabular}{l l c c c c c c c c }
\hline
{\begin{tabular}{c} Dataset \\ {\small [output dim]} \end{tabular}} & Metric & Mahalanobis & HPD & ECM & OT & MVCS & PCP & L-CP & C-PCP \\
\hline
\multirow{4}{*}{{Bias [2]}} & ERT [\%] & $\mathbf{1.30_{0.22}}$ & $1.67_{0.23}$ & $2.25_{0.20}$ & $2.14_{0.34}$ & $1.54_{0.22}$ & $\underline{1.43_{0.35}}$ & $1.58_{0.30}$ & $1.66_{0.46}$ \\
 & WSC & $0.75_{0.01}$ & $\underline{0.76_{0.02}}$ & $0.75_{0.01}$ & $0.76_{0.02}$ & $0.74_{0.01}$ & $0.75_{0.01}$ & $0.75_{0.01}$ & $\mathbf{0.78_{0.03}}$ \\
 & Volume & $\mathbf{1.10_{0.02}}$ & $1.68_{0.04}$ & $1.33_{0.04}$ & $1.34_{0.05}$ & $\underline{1.30_{0.04}}$ & $1.54_{0.02}$ & $1.66_{0.04}$ & $1.79_{0.06}$ \\
 & Coverage & $0.95_{0.00}$ & $0.96_{0.01}$ & $0.95_{0.00}$ & $0.96_{0.00}$ & $0.95_{0.00}$ & $0.95_{0.00}$ & $0.95_{0.00}$ & $0.96_{0.01}$ \\
 & Time [s] & $\underline{0.02_{0.00}}$ & $808.27_{150.00}$ & $\mathbf{0.01_{0.00}}$ & $17.04_{4.22}$ & $0.14_{0.03}$ & $361.15_{43.23}$ & $3.31_{0.03}$ & $1779.19_{430.56}$ \\
\hline
\multirow{4}{*}{{CASP [2]}} & ERT [\%] & $\mathbf{1.27_{0.14}}$ & $2.16_{0.12}$ & $4.86_{0.15}$ & $4.24_{0.13}$ & $2.31_{0.12}$ & $2.52_{0.13}$ & $3.47_{0.13}$ & $\underline{2.07_{0.12}}$ \\
 & WSC & $\mathbf{0.87_{0.00}}$ & $\underline{0.82_{0.01}}$ & $0.65_{0.01}$ & $0.75_{0.01}$ & $0.78_{0.01}$ & $0.77_{0.01}$ & $0.79_{0.01}$ & $0.81_{0.01}$ \\
 & Volume & $\mathbf{1.12_{0.01}}$ & $1.74_{0.02}$ & $1.60_{0.01}$ & $2.30_{0.48}$ & $\underline{1.12_{0.01}}$ & $1.78_{0.01}$ & $1.85_{0.03}$ & $1.92_{0.03}$ \\
 & Coverage & $0.95_{0.00}$ & $0.95_{0.00}$ & $0.95_{0.00}$ & $0.95_{0.00}$ & $0.95_{0.00}$ & $0.95_{0.00}$ & $0.95_{0.00}$ & $0.95_{0.00}$ \\
 & Time [s] & $\underline{0.01_{0.00}}$ & $679.71_{76.49}$ & $\mathbf{0.00_{0.00}}$ & $0.55_{0.08}$ & $0.09_{0.01}$ & $415.35_{58.90}$ & $3.17_{0.14}$ & $1736.05_{19.29}$ \\
\hline
\multirow{4}{*}{{House [2]}} & ERT [\%] & $\underline{1.80_{0.21}}$ & $1.89_{0.20}$ & $4.18_{0.22}$ & $3.78_{0.16}$ & $3.80_{0.11}$ & $2.90_{0.19}$ & $1.90_{0.19}$ & $\mathbf{1.71_{0.20}}$ \\
 & WSC & $0.83_{0.01}$ & $\mathbf{0.83_{0.01}}$ & $0.75_{0.01}$ & $0.75_{0.01}$ & $0.78_{0.01}$ & $0.80_{0.01}$ & $0.82_{0.00}$ & $\underline{0.83_{0.00}}$ \\
 & Volume & $\mathbf{0.93_{0.01}}$ & $1.46_{0.02}$ & $1.52_{0.03}$ & $2.34_{0.12}$ & $\underline{1.26_{0.05}}$ & $1.44_{0.02}$ & $1.52_{0.01}$ & $1.56_{0.02}$ \\
 & Coverage & $0.95_{0.00}$ & $0.96_{0.00}$ & $0.95_{0.00}$ & $0.95_{0.00}$ & $0.95_{0.00}$ & $0.95_{0.00}$ & $0.95_{0.00}$ & $0.95_{0.00}$ \\
 & Time [s] & $\underline{0.01_{0.00}}$ & $654.83_{142.51}$ & $\mathbf{0.00_{0.00}}$ & $0.21_{0.01}$ & $0.07_{0.00}$ & $411.61_{45.24}$ & $3.35_{0.11}$ & $1580.50_{23.19}$ \\
\hline
\multirow{4}{*}{{rf1 [8]}} & ERT [\%] & $\mathbf{1.82_{0.15}}$ & N/A & $6.86_{0.22}$ & $\underline{1.95_{0.21}}$ & $4.68_{0.24}$ & $5.20_{0.28}$ & $2.58_{0.34}$ & $2.86_{0.36}$ \\
 & WSC & $\mathbf{0.75_{0.01}}$ & N/A & $0.53_{0.03}$ & $\underline{0.75_{0.02}}$ & $0.63_{0.03}$ & $0.66_{0.02}$ & $0.69_{0.03}$ & $0.72_{0.02}$ \\
 & Volume & $\mathbf{0.31_{0.01}}$ & N/A & $1.32_{0.03}$ & $0.56_{0.02}$ & $\underline{0.36_{0.01}}$ & $0.52_{0.00}$ & $0.53_{0.01}$ & $0.52_{0.00}$ \\
 & Coverage & $0.95_{0.00}$ & N/A & $0.95_{0.00}$ & $0.96_{0.00}$ & $0.95_{0.00}$ & $0.95_{0.00}$ & $0.95_{0.00}$ & $0.95_{0.00}$ \\
 & Time [s] & $\underline{0.05_{0.01}}$ & N/A & $\mathbf{0.01_{0.00}}$ & $16.44_{3.72}$ & $0.14_{0.03}$ & $414.11_{23.15}$ & $2.86_{0.17}$ & $1588.91_{96.69}$ \\
\hline
\multirow{4}{*}{{rf2 [8]}} & ERT [\%] & $\mathbf{2.14_{0.39}}$ & N/A & $6.35_{0.29}$ & $3.25_{0.33}$ & $4.59_{0.38}$ & $3.96_{0.25}$ & $\underline{2.19_{0.34}}$ & $2.50_{0.25}$ \\
 & WSC & $0.72_{0.01}$ & N/A & $0.72_{0.01}$ & $\underline{0.76_{0.02}}$ & $0.71_{0.02}$ & $0.73_{0.01}$ & $0.71_{0.01}$ & $\mathbf{0.77_{0.01}}$ \\
 & Volume & $\mathbf{0.39_{0.02}}$ & N/A & $1.39_{0.03}$ & $1.19_{0.06}$ & $0.51_{0.02}$ & $\underline{0.47_{0.01}}$ & $0.48_{0.01}$ & $0.48_{0.00}$ \\
 & Coverage & $0.95_{0.00}$ & N/A & $0.95_{0.00}$ & $0.96_{0.00}$ & $0.95_{0.00}$ & $0.95_{0.00}$ & $0.95_{0.00}$ & $0.96_{0.00}$ \\
 & Time [s] & $\underline{0.03_{0.00}}$ & N/A & $\mathbf{0.01_{0.00}}$ & $35.98_{5.25}$ & $0.08_{0.00}$ & $352.59_{19.86}$ & $3.34_{0.63}$ & $1551.95_{78.70}$ \\
\hline
\multirow{4}{*}{{scm1d [16]}} & ERT [\%] & $\mathbf{1.53_{0.21}}$ & N/A & $6.42_{0.20}$ & $2.15_{0.35}$ & $6.32_{0.22}$ & $6.86_{0.19}$ & $\underline{2.14_{0.24}}$ & $3.87_{0.58}$ \\
 & WSC & $0.75_{0.01}$ & N/A & $0.54_{0.02}$ & $\underline{0.76_{0.03}}$ & $0.55_{0.02}$ & $0.54_{0.02}$ & $0.74_{0.01}$ & $\mathbf{0.92_{0.04}}$ \\
 & Volume & $1.27_{0.01}$ & N/A & $2.12_{0.04}$ & $1.72_{0.03}$ & $1.69_{0.04}$ & $\mathbf{1.13_{0.01}}$ & $1.28_{0.01}$ & $\underline{1.26_{0.01}}$ \\
 & Coverage & $0.95_{0.00}$ & N/A & $0.95_{0.00}$ & $0.95_{0.01}$ & $0.95_{0.00}$ & $0.95_{0.00}$ & $0.95_{0.00}$ & $0.98_{0.01}$ \\
 & Time [s] & $0.08_{0.01}$ & N/A & $\mathbf{0.01_{0.00}}$ & $28.33_{7.19}$ & $\underline{0.08_{0.00}}$ & $361.67_{19.17}$ & $2.78_{0.19}$ & $1720.17_{194.29}$ \\
\hline
\multirow{4}{*}{{scm20d [16]}} & ERT [\%] & $\mathbf{1.16_{0.24}}$ & N/A & $6.49_{0.25}$ & $\underline{1.84_{0.21}}$ & $5.84_{0.30}$ & $6.54_{0.23}$ & $1.88_{0.32}$ & $4.17_{0.26}$ \\
 & WSC & $0.76_{0.01}$ & N/A & $0.67_{0.01}$ & $\underline{0.76_{0.01}}$ & $0.70_{0.01}$ & $0.68_{0.01}$ & $\mathbf{0.77_{0.01}}$ & $0.75_{0.02}$ \\
 & Volume & $\underline{1.44_{0.02}}$ & N/A & $2.33_{0.05}$ & $2.66_{0.08}$ & $\mathbf{1.22_{0.06}}$ & $1.60_{0.01}$ & $1.91_{0.02}$ & $1.73_{0.02}$ \\
 & Coverage & $0.95_{0.00}$ & N/A & $0.95_{0.00}$ & $0.96_{0.00}$ & $0.95_{0.00}$ & $0.95_{0.00}$ & $0.96_{0.00}$ & $0.96_{0.00}$ \\
 & Time [s] & $\underline{0.09_{0.01}}$ & N/A & $\mathbf{0.01_{0.00}}$ & $21.72_{5.12}$ & $0.11_{0.02}$ & $393.40_{23.88}$ & $2.82_{0.20}$ & $1839.43_{405.44}$ \\
\hline
\multirow{4}{*}{{Taxi [2]}} & ERT [\%] & $5.29_{0.12}$ & $\mathbf{4.26_{0.10}}$ & $7.11_{0.07}$ & $6.58_{0.11}$ & $6.89_{0.08}$ & $4.84_{0.09}$ & $5.34_{0.10}$ & $\underline{4.34_{0.12}}$ \\
 & WSC & $0.71_{0.01}$ & $\underline{0.74_{0.01}}$ & $0.53_{0.01}$ & $0.63_{0.01}$ & $0.53_{0.01}$ & $0.71_{0.01}$ & $0.70_{0.01}$ & $\mathbf{0.75_{0.01}}$ \\
 & Volume & $\underline{3.36_{0.01}}$ & $\mathbf{3.24_{0.06}}$ & $4.14_{0.02}$ & $3.45_{0.37}$ & $3.93_{0.03}$ & $3.71_{0.01}$ & $3.45_{0.01}$ & $3.84_{0.03}$ \\
 & Coverage & $0.95_{0.00}$ & $0.95_{0.00}$ & $0.95_{0.00}$ & $0.95_{0.00}$ & $0.95_{0.00}$ & $0.95_{0.00}$ & $0.95_{0.00}$ & $0.95_{0.00}$ \\
 & Time [s] & $\underline{0.01_{0.00}}$ & $400.28_{10.44}$ & $\mathbf{0.00_{0.00}}$ & $0.74_{0.11}$ & $0.09_{0.02}$ & $382.11_{49.91}$ & $3.03_{0.18}$ & $1611.38_{12.59}$ \\
\hline
\end{tabular}
}
\label{tab:app:results:revealed:alpha005}
\end{table}

\subsection{Projection of the outputs}
\label{app:exp:projection}

Here, we consider the conformal sets constructed for a low-rank linear projection of the output $MY$ with $M \in \rb^{p \times k}$ randomly generated.
For 10 different seeds, we randomly generate the projection matrix $M$ and evaluate the average volume and marginal coverage achieved by the strategy described in Section~\ref{sec:projection:output}. While most strategies like HPD, C-PCP, L-CP, PCP and MVCS do not generalize to projection of the output without prohibitive computational time, the other baselines can be extended by directly conformalizing $MY$.
We report results in Tables~\ref{tab:app:results:projection:alpha01}~and~\ref{tab:app:results:projection:alpha005} for a coverage level of $1-\alpha = 0.9$, and $1-\alpha = 0.95$ respectively.

Once again the conformal sets constructed using the projected density distribution $\hat{p}(MY|X)$ generally achieve smaller size than the projected conformal sets.

\begin{table}[ht]
\centering
\small
\caption{Comparison of conditional coverage ($\alpha=0.1$) for a projection of the outputs for different conformal methods using the ERT metric (lower is better) and WSC (closer to $0.9$ is better). Best values in bold, second best underlined. N/A indicates that the method failed to produce valid results for the corresponding dataset due to poor conditional density estimation leading to numerical issues in high dimensions. Experiments are repeated 10 times, and the index number is the standard error across those 10 experiments.}
\begin{tabular}{l l c c c }
\hline
Dataset & Metric & Mahalanobis & ECM & OT \\
\hline
\multirow{4}{*}{{Bias}} & ERT [\%] & $\mathbf{2.62_{0.39}}$ & $4.70_{0.49}$ & $\underline{3.64_{0.50}}$ \\
 & WSC & $\mathbf{0.73_{0.01}}$ & $0.72_{0.01}$ & $\underline{0.73_{0.01}}$ \\
 & Volume & $\mathbf{1.34_{0.15}}$ & $\underline{1.52_{0.16}}$ & $2.00_{0.22}$ \\
 & Coverage & $0.90_{0.01}$ & $0.90_{0.01}$ & $0.90_{0.01}$ \\
\hline
\multirow{4}{*}{{CASP}} & ERT [\%] & $\mathbf{2.04_{0.28}}$ & $5.03_{0.16}$ & $\underline{4.80_{0.25}}$ \\
 & WSC & $\mathbf{0.82_{0.00}}$ & $\underline{0.79_{0.01}}$ & $0.78_{0.01}$ \\
 & Volume & $\mathbf{1.90_{0.23}}$ & $\underline{2.16_{0.26}}$ & $2.18_{0.27}$ \\
 & Coverage & $0.90_{0.00}$ & $0.90_{0.00}$ & $0.90_{0.00}$ \\
\hline
\multirow{4}{*}{{House}} & ERT [\%] & $\mathbf{2.99_{0.31}}$ & $7.87_{0.37}$ & $\underline{7.59_{0.32}}$ \\
 & WSC & $\mathbf{0.77_{0.00}}$ & $0.73_{0.01}$ & $\underline{0.74_{0.01}}$ \\
 & Volume & $\mathbf{1.92_{0.17}}$ & $\underline{2.59_{0.24}}$ & $2.64_{0.25}$ \\
 & Coverage & $0.90_{0.00}$ & $0.90_{0.00}$ & $0.90_{0.00}$ \\
\hline
\multirow{4}{*}{{rf1}} & ERT [\%] & $\mathbf{3.81_{0.67}}$ & $8.88_{0.64}$ & $\underline{5.17_{0.36}}$ \\
 & WSC & $\mathbf{0.73_{0.01}}$ & $0.61_{0.02}$ & $\underline{0.73_{0.01}}$ \\
 & Volume & $\mathbf{0.93_{0.08}}$ & $\underline{1.30_{0.13}}$ & $1.73_{0.11}$ \\
 & Coverage & $0.90_{0.00}$ & $0.90_{0.00}$ & $0.91_{0.00}$ \\
\hline
\multirow{4}{*}{{rf2}} & ERT [\%] & $\mathbf{4.38_{0.37}}$ & $8.33_{0.54}$ & $\underline{5.94_{0.37}}$ \\
 & WSC & $\underline{0.73_{0.01}}$ & $0.72_{0.01}$ & $\mathbf{0.73_{0.01}}$ \\
 & Volume & $\mathbf{1.17_{0.17}}$ & $\underline{1.55_{0.21}}$ & $2.35_{0.22}$ \\
 & Coverage & $0.90_{0.00}$ & $0.90_{0.00}$ & $0.90_{0.01}$ \\
\hline
\multirow{4}{*}{{scm1d}} & ERT [\%] & $\mathbf{1.97_{0.45}}$ & $9.21_{0.41}$ & $\underline{5.92_{0.39}}$ \\
 & WSC & $\mathbf{0.75_{0.01}}$ & $0.64_{0.01}$ & $\underline{0.72_{0.02}}$ \\
 & Volume & $\mathbf{4.31_{0.18}}$ & $\underline{4.82_{0.16}}$ & $5.33_{0.21}$ \\
 & Coverage & $0.90_{0.01}$ & $0.90_{0.00}$ & $0.90_{0.01}$ \\
\hline
\multirow{4}{*}{{scm20d}} & ERT [\%] & $\mathbf{2.33_{0.46}}$ & $7.29_{0.43}$ & $\underline{4.42_{0.39}}$ \\
 & WSC & $\underline{0.73_{0.01}}$ & $0.71_{0.01}$ & $\mathbf{0.75_{0.02}}$ \\
 & Volume & $\mathbf{5.63_{0.31}}$ & $\underline{6.59_{0.35}}$ & $8.31_{0.44}$ \\
 & Coverage & $0.89_{0.01}$ & $0.90_{0.00}$ & $0.91_{0.01}$ \\
\hline
\multirow{4}{*}{{Taxi}} & ERT [\%] & $\mathbf{2.09_{0.26}}$ & $6.00_{0.95}$ & $\underline{5.49_{0.59}}$ \\
 & WSC & $\mathbf{0.83_{0.01}}$ & $0.74_{0.03}$ & $\underline{0.76_{0.01}}$ \\
 & Volume & $\mathbf{3.71_{0.48}}$ & $4.17_{0.42}$ & $\underline{4.09_{0.45}}$ \\
 & Coverage & $0.90_{0.00}$ & $0.90_{0.00}$ & $0.90_{0.00}$ \\
\hline
\end{tabular}
\label{tab:app:results:projection:alpha01}
\end{table}

\begin{table}[ht]
\centering
\small
\caption{Comparison of conditional coverage ($\alpha=0.05$) for a projection of the outputs for different conformal methods using the ERT metric (lower is better) and WSC (closer to $0.95$ is better). Best values in bold, second best underlined. N/A indicates that the method failed to produce valid results for the corresponding dataset due to poor conditional density estimation leading to numerical issues in high dimensions. Experiments are repeated 10 times, and the index number is the standard error across those 10 experiments.}
\begin{tabular}{l l c c c }
\hline
Dataset & Metric & Mahalanobis & ECM & OT \\
\hline
\multirow{4}{*}{{Bias}} & ERT [\%] & $\mathbf{0.93_{0.29}}$ & $\underline{2.20_{0.29}}$ & $2.35_{0.40}$ \\
 & WSC & $0.74_{0.01}$ & $\underline{0.75_{0.01}}$ & $\mathbf{0.77_{0.02}}$ \\
 & Volume & $\mathbf{1.64_{0.18}}$ & $\underline{1.96_{0.20}}$ & $2.47_{0.26}$ \\
 & Coverage & $0.95_{0.00}$ & $0.95_{0.00}$ & $0.96_{0.01}$ \\
\hline
\multirow{4}{*}{{CASP}} & ERT [\%] & $\mathbf{0.82_{0.17}}$ & $2.72_{0.15}$ & $\underline{2.39_{0.14}}$ \\
 & WSC & $\mathbf{0.87_{0.00}}$ & $0.82_{0.01}$ & $\underline{0.84_{0.01}}$ \\
 & Volume & $\mathbf{2.33_{0.28}}$ & $\underline{2.62_{0.31}}$ & $2.91_{0.30}$ \\
 & Coverage & $0.95_{0.00}$ & $0.95_{0.00}$ & $0.95_{0.00}$ \\
\hline
\multirow{4}{*}{{House}} & ERT [\%] & $\mathbf{1.87_{0.15}}$ & $4.28_{0.31}$ & $\underline{3.62_{0.33}}$ \\
 & WSC & $\mathbf{0.80_{0.01}}$ & $0.76_{0.01}$ & $\underline{0.78_{0.01}}$ \\
 & Volume & $\mathbf{2.34_{0.21}}$ & $\underline{3.23_{0.32}}$ & $3.67_{0.35}$ \\
 & Coverage & $0.95_{0.00}$ & $0.95_{0.00}$ & $0.95_{0.00}$ \\
\hline
\multirow{4}{*}{{rf1}} & ERT [\%] & $\mathbf{1.81_{0.42}}$ & $5.72_{0.30}$ & $\underline{2.70_{0.39}}$ \\
 & WSC & $\underline{0.75_{0.01}}$ & $0.59_{0.03}$ & $\mathbf{0.77_{0.01}}$ \\
 & Volume & $\mathbf{1.08_{0.09}}$ & $\underline{1.94_{0.18}}$ & $2.13_{0.14}$ \\
 & Coverage & $0.95_{0.00}$ & $0.95_{0.00}$ & $0.96_{0.00}$ \\
\hline
\multirow{4}{*}{{rf2}} & ERT [\%] & $\mathbf{1.82_{0.40}}$ & $5.08_{0.35}$ & $\underline{3.28_{0.30}}$ \\
 & WSC & $\underline{0.75_{0.01}}$ & $0.73_{0.01}$ & $\mathbf{0.77_{0.02}}$ \\
 & Volume & $\mathbf{1.37_{0.19}}$ & $\underline{2.10_{0.28}}$ & $3.15_{0.31}$ \\
 & Coverage & $0.95_{0.00}$ & $0.95_{0.00}$ & $0.96_{0.01}$ \\
\hline
\multirow{4}{*}{{scm1d}} & ERT [\%] & $\mathbf{0.91_{0.40}}$ & $5.38_{0.22}$ & $\underline{3.43_{0.21}}$ \\
 & WSC & $\mathbf{0.78_{0.01}}$ & $0.60_{0.02}$ & $\underline{0.75_{0.02}}$ \\
 & Volume & $\mathbf{5.20_{0.22}}$ & $6.28_{0.21}$ & $\underline{6.15_{0.35}}$ \\
 & Coverage & $0.95_{0.00}$ & $0.95_{0.00}$ & $0.96_{0.00}$ \\
\hline
\multirow{4}{*}{{scm20d}} & ERT [\%] & $\mathbf{1.34_{0.33}}$ & $4.91_{0.25}$ & $\underline{2.35_{0.33}}$ \\
 & WSC & $\underline{0.77_{0.01}}$ & $0.75_{0.01}$ & $\mathbf{0.79_{0.02}}$ \\
 & Volume & $\mathbf{6.67_{0.37}}$ & $\underline{8.53_{0.46}}$ & $9.91_{0.72}$ \\
 & Coverage & $0.95_{0.00}$ & $0.96_{0.00}$ & $0.96_{0.01}$ \\
\hline
\multirow{4}{*}{{Taxi}} & ERT [\%] & $\mathbf{1.22_{0.12}}$ & $3.76_{0.58}$ & $\underline{3.27_{0.27}}$ \\
 & WSC & $\mathbf{0.87_{0.01}}$ & $0.75_{0.03}$ & $\underline{0.78_{0.01}}$ \\
 & Volume & $\mathbf{4.49_{0.59}}$ & $5.13_{0.48}$ & $\underline{4.98_{0.48}}$ \\
 & Coverage & $0.95_{0.00}$ & $0.95_{0.00}$ & $0.95_{0.00}$ \\
\hline
\end{tabular}
\label{tab:app:results:projection:alpha005}
\end{table}

\section{Details on the datasets}
\label{app:dataset:information}
See Table~\ref{tab:dataset:information}. For the experimental setup with missing values and projection of the outputs, we added one feature to the output vector when the dimension of the output is two. 

\begin{table}[h!]
\centering
\caption{Description of the datasets.}
\begin{tabular}{@{}lcccc@{}}
\toprule
Dataset & \shortstack{Number of \\ samples} & \shortstack{Dimension \\ of feature} & \shortstack{Dimension \\ of response} & \shortstack{Number of \\ revealed outputs} \\
\midrule
Bias \citep{cho2020comparative} & 7752 & 22 & 2 & 1 \\ \hline
CASP \citep{rana2013physicochemical} & 45730 & 8 & 2  & 1 \\ \hline
House \citep{pace1997sparse} & 21613 & 17 & 2 & 1 \\ \hline
rf1 \citep{tsoumakas2011mulan} & 9125 & 64 & 8 & 5 \\ \hline
rf2 \citep{tsoumakas2011mulan} & 9125 & 576 & 8 & 5 \\ \hline
scm1d \citep{tsoumakas2011mulan} & 9803 & 280 & 16 & 5 \\ \hline
scm20d \citep{tsoumakas2011mulan} & 8966 & 61 & 16 & 5 \\ \hline
Taxi \citep{wang2023conformal} & 61286 & 6 & 2 & 1 \\ \hline
\bottomrule
\end{tabular}
\label{tab:dataset:information}
\end{table}

\section{Hyperparameters}
\label{sec:hyperparameters}

Each dataset requires its own hyperparameter tuning. In Table~\ref{tab:hyperparameters}, we show the hyperparameters used for the CASP dataset. Full details for all experiments can be found in the attached ZIP file. All models have been trained on a CPU device. 

\begin{table}[t]
\centering
\caption{Hyper-parameters on the CASP dataset.}
\begin{tabular}{@{}lcccc@{}}
\toprule
Hyper-parameters & Value \\
\midrule
Hidden dimension (center) & 256 \\
Number of hidden layers (center) & 1 \\
Activation function (center) & ReLU \\
Hidden dimension (matrix) & 256 \\
Number of hidden layers (matrix) & 0 \\
Activation function (matrix) & ReLU \\
Optimizer & Adam \\
Number of epochs & 250 \\
Batch size & 100 \\
Learning rate (model) & 0.001 \\
Learning rate (matrix) & 0.001 \\
Learning scheduler & CosineAnnealingLR \\
\bottomrule
\end{tabular}
\label{tab:hyperparameters}
\end{table}

\end{appendices}

\end{document}